\documentclass[10pt,american,english,a4paper,twocolumn]{article}
\usepackage[affil-it]{authblk} 
\usepackage[T1]{fontenc}
\usepackage[utf8]{inputenc}
\usepackage{babel}
\usepackage{array}
\usepackage{verbatim}
\usepackage{refstyle}
\usepackage{amsthm}
\usepackage{amsmath}
\usepackage{graphicx}
\usepackage[unicode=true]{hyperref}
\makeatletter

\numberwithin{equation}{section}
\numberwithin{figure}{section}

\usepackage{multicol} 
\usepackage{xcolor} 
\PassOptionsToPackage {hyphens}{url} 
\usepackage{hyperref}
\PassOptionsToPackage{pdftex}{graphicx}
\usepackage{graphicx}
\usepackage{eurosym}

\usepackage[hmargin=1.4cm,vmargin=1.8cm]{geometry}

\PassOptionsToPackage{dvipsnames}{xcolor}
\RequirePackage{xcolor} 
\definecolor{halfgray}{gray}{0.55} 
\definecolor{webgreen}{rgb}{0,.5,0}
\definecolor{webbrown}{rgb}{.6,0,0}
\definecolor{Maroon}{cmyk}{0, 0.87, 0.68, 0.32}
\definecolor{RoyalBlue}{cmyk}{1, 0.50, 0, 0}
\definecolor{Lavender}{cmyk}{0,0.48,0,0}
\definecolor{CarnationPink}{cmyk}{0,0.63,0,0} 
\definecolor{Black}{cmyk}{0, 0, 0, 0}

\hypersetup{
    colorlinks=true, linktocpage=true, pdfstartpage=1, pdfstartview=FitV,
    breaklinks=true, pdfpagemode=UseNone, pageanchor=true, pdfpagemode=UseOutlines,%
    plainpages=false, bookmarksnumbered, bookmarksopen=true, bookmarksopenlevel=1,%
    hypertexnames=true, pdfhighlight=/O,
    urlcolor=CarnationPink, linkcolor=webgreen ,citecolor=RoyalBlue, pagecolor=RoyalBlue,%
}

\usepackage{prettyref}
\newrefformat{tab}{Table\,\ref{#1}}
\newrefformat{fig}{Figure\,\ref{#1}}
\newrefformat{eq}{Eq.\,\textup{(\ref{#1})}}

\PassOptionsToPackage{square,numbers}{natbib}

\usepackage[maxfloats=71]{morefloats}

\usepackage[font=scriptsize,labelfont=bf]{caption}

\let\OLDthebibliography\thebibliography
\renewcommand\thebibliography[1]{
  \OLDthebibliography{#1}
  \setlength{\parskip}{0pt plus 0.2ex}
  \setlength{\itemsep}{1pt plus 31.8ex}
}

\makeatother

\begin{document}

\renewcommand\Affilfont{\itshape\footnotesize}
\title{Deep Learning Convolutional Networks for Multiphoton Microscopy Vasculature Segmentation}

\author[a]{Petteri Teikari%
  \thanks{E-mail: \texttt{petteri.teikari@gmail.com}; Corresponding author}}
\author[a]{Marc Santos}
\author[a,b]{Charissa Poon}
\author[a,c]{Kullervo Hynynen}

\affil[a]{Physical Sciences, Sunnybrook Research Institute, Toronto, Canada}
\affil[b]{Institute of Biomaterials and Biomedical Engineering, University of Toronto, Canada}
\affil[c]{Department of Medical Biophysics, University of Toronto, Toronto Medical Discovery Tower, Canada}
\maketitle


\begin{abstract}
Recently there has been an increasing trend to use deep learning frameworks
for both 2D consumer images and for 3D medical images. However, there
has been little effort to use deep frameworks for volumetric vascular
segmentation. We wanted to address this by providing a freely available
dataset of 12 annotated two-photon vasculature microscopy stacks.
We demonstrated the use of deep learning framework consisting both
2D and 3D convolutional filters (ConvNet). Our hybrid 2D-3D architecture
produced promising segmentation result. We derived the architectures
from Lee \emph{et al. }who used the ZNN framework initially designed
for electron microscope image segmentation. We hope that by sharing
our volumetric vasculature datasets, we will inspire other researchers
to experiment with vasculature dataset and improve the used network
architectures.
\end{abstract}
\maketitle

\tableofcontents{}

\section{Introduction}

Quantitative analysis of brain vasculature is used in a variety of
fields, including vascular development \cite{connor2015anintegrated,segarra2015avascular,newberry2015testing}
and physiology \cite{shih2015robustand}, neurovascular coupling \cite{tsai2009correlations,dalkara2015cerebral},
and blood-brain barrier studies \cite{nhan2013drugdelivery,burgess2014analysis}.
Distinguishing blood vessels from the surrounding tissue (vessel segmentation)
is often a necessary preliminary step that enables more accurate and
efficient analyses of the vascular network. For example, characteristics
of vasculature morphology such as tortuosity, length, and diameter,
can be obtained without confounding factors from the extravascular
space, such as dendrites. In addition to characterizing the vasculature
itself, vessel segmentation also facilitates analyses of other dynamic
factors, including cortical blood flow and angiogenesis. 

Clinically, quantitative analysis of vessels will assist in making
diagnoses and planning surgeries \cite{lesage2009areview,rudyanto2014comparing,yun2015stenosis}.
For example, retinal vasculature imaging \cite{pinhas2013invivo,spaide2015retinal}
allows inexpensive and fast screening of several eye-related and systematic
pathologies such as glaucoma, age-related macular degeneration, diabetic
retinopathy, hypertension, arteriosclerosis and Alzheimer's disease
\cite{ikram2013retinal}. Differentiating blood vessels from the
surrounding tissue also allows more accurate analyses of extravascular
structures, such as tumor volume quantification \cite{reeves2006onmeasuring}
and pulmonary lobes structural analysis \cite{lassen2013automatic}.
Given that vascular diseases, such as coronary heart disease, are
among the largest public health problems in developed countries \cite{worldhealthorganization2008thetop},
accurate and efficient image analysis will only become more relevant.
\cite{worldhealthorganization2008thetop}. Thus, the segmentation
of vascular structures from surrounding tissue is useful for both
basic research and clinical applications. There have been various
approaches for vessel segmentation (for reviews see \cite{kirbas2004areview,lesage2009areview}),
but to date, no single method have been able to successfully segment
vessels from every imaging modality and every organ \cite{rudyanto2014comparing}. 

Our group uses vessel segmentation for two purposes: 1) To analyze
changes in vascular morphology after focused ultrasound mediated blood-brain
barrier opening \cite{hynynen2005localand,burgess2016microbubbleassisted},
and 2) to observe tumor pathophysiology and drug kinetics following
application of focused ultrasound stimulated microbubbles (unpublished).
Both of these projects use the two-photon microscope for acquiring
high-resolution images. We were motivated to improve our vessel segmentation
pipelines from previous custom-written semi-automatic Matlab scripts
\cite{nhan2013drugdelivery}, and labor-intensive manual approaches
using commercial Imaris (Bitplane AG, Zurich, Switzerland) platform
and open-source ImageJ/FIJI platform \cite{schindelin2015theimagej},
to be more automatic and robust. Delineating blood vessels from the
extravascular space enables quantification of the rate and duration
of dye leakage, which can be correlated with kinetics and characteristics
of blood-brain barrier integrity \cite{cho2011twophoton,nhan2013drugdelivery,burgess2014analysis}.
Other groups have used vessel segmentation as an image processing
tool to analyze other factors from two-photon datasets, including
neurovascular coupling \cite{tran2015acutetwophoton}, neuronal calcium
imaging \cite{daniel2015optical,maeda2015weaksinusoidal}, and low-intensity
focused ultrasound brain modulation paradigms \cite{tufail2011ultrasonic,bystritsky2015areview,moore2015manipulating}. 

Two-photon microscopy, or more generally, multiphoton microscopy,
has become the workhorse of neuronal imaging \cite{helmchen2005deeptissue}.
Multiphoton microscopy allows better optical sectioning and reduced
photobleaching outside of the imaging plane compared to the traditional
confocal techniques due to the nonlinear nature of the two-photon
excitation fluorescence. Traditional two-photon microscopy operates
on scanning point-by-point compared to whole-field approach of confocal
microscopy, limiting also the maximum frame rates achieved by scanning
two-photon microscopy. Two-photon light-sheet imaging operates on
a line or a plane basis instead of a point, speeding the volumetric
imaging by one or two orders of magnitude if faster faster rates are
needed \cite{truong2011deepand}. Additionally two-photon fluorescence
imaging can be combined with other nonlinear processes such as with
third-harmonic generation (THG) for label-free vascular imaging \cite{witte2011labelfree},
and other microscopy techniques such as electron microscopy for more
detailed analysis \cite{bishop2011nearinfrared}. Silvestri \emph{et
al.} \cite{silvestri2014correlative} for example integrate \emph{in
vivo }two-photon microscopy with \emph{ex vivo }light sheet microscopy
and use the major blood vessels as landmark points for registration.

Compared to the literature focused on clinical angiography with various
modalities and anatomical applications, there exists very little literature
devoted on processing multiphoton vasculature images. Likewise, not
much work has been done on open-source software and/or code for multiphoton
vasculature analysis. The work by Santamaria-Pang \emph{et al.} \cite{santamaria-pang2015automatic}
on tubular 3D neuronal structures representing one of the few examples
for ``morphological'' multiphoton microscopy analysis, and Python-based
VMTK (\cite{antiga2012vmtkvascular}, \href{http://www.vmtk.org/}{http://www.vmtk.org/})
for open-source vessel analysis. This is in stark contrast to work
devoted on calcium imaging analysis with various freely available
toolboxes (e.g. \cite{mukamel2009automated,tomek2013twophoton,muir2015focusstack,patel2015automated}).

Traditionally vessel segmentation have been done on some combination
of vascular models, image features and extraction schemes often relying
on prior knowledge about the tubularity of vessels \cite{kirbas2004areview,lesage2009areview}.
Typically in computer vision/image analysis field, algorithms and
pipelines are developed using reference dataset as benchmarks for
performance. In biomedical image analysis, almost all open image segmentation
challenges are listed in \href{ http://grand-challenge.org/}{ http://grand-challenge.org/}
with only challenge (VESSEL12, \cite{rudyanto2014comparing}) devoted
to vessel segmentation. It is common that many fields suffer from
lack of annotated datasets \cite{ferguson2014bigdata} as they are
expensive to generate such as is the case for example in high content
screening (HCS) technologies labeled at cell level \cite{kraus2015classifying,ljosa2012annotated},
and in electron microscopy \cite{arganda-carreras2015crowdsourcing}.
Additional standardized datasets can be found for evaluating coronary
artery centerline extraction algorithms \cite{schaap2009standardized},
and for evaluating coronary artery stenosis detection, stenosis quantification
and lumen segmentation algorithms in computed tomography angiography
\cite{kiricsli2013standardized}.

Thus, despite the numerous papers on vessel segmentation there has
been very little effort for creating standardized three-dimensional
vascular datasets. The the most similar datasets can be found for
example for two-dimension retinal vessels in DRIVE dataset \cite{staal2004ridgebased},
and three-dimension tubular fibers in DIADEM challenge \cite{brown2011thediadem}.
Among the 23 submitted methods to the VESSEL12 challenge, only two
submission were machine-learning based with the other one of them
ending up providing the best overall performance in terms of segmentation
accuracy. Similarly with natural images, research teams compete against
each other trying to improve the performance of the classifier. One
example of such challenge is the ImageNet Large Scale Visual Recognition
Challenge (ILSVRC) challenge that is taking place annually with the
same database of images \cite{russakovsky2014imagenet}. 

During past few years, data-driven machine learning algorithms have
replaced ``hand-crafted'' filter pipelines on many fields of image
processing. Majority of the emerged approaches have relied on deep
learning networks \cite{lecun2015deeplearning,schmidhuber2015deeplearning,lake2015humanlevel}
opposed to ``traditional'' shallow networks \cite{bianchini2014onthe}.
\textbf{\emph{}}From different deep learning architectures, convolutional
neural networks (CNN or ConvNet) have been the mostly used in image
classification and image segmentation. While ConvNets have been around
for decades (e.g. \cite{ivakhnenko1971polynomial,lecun1989backpropagation}),
the recent success had been due to the combination of bigger annotated
datasets, more powerful hardware, new ideas, algorithms and improved
network architectures enabling this sort of ``paradigm shift'' in
machine learning. Since 2011, graphical processing unit (GPU)-based
ConvNets have dominated classification (\cite{krizhevsky2012imagenet}),
and segmentation contests (\cite{ciresan2012deepneural}).

\begin{figure*}
\centerline{\includegraphics[width=1.8\columnwidth]{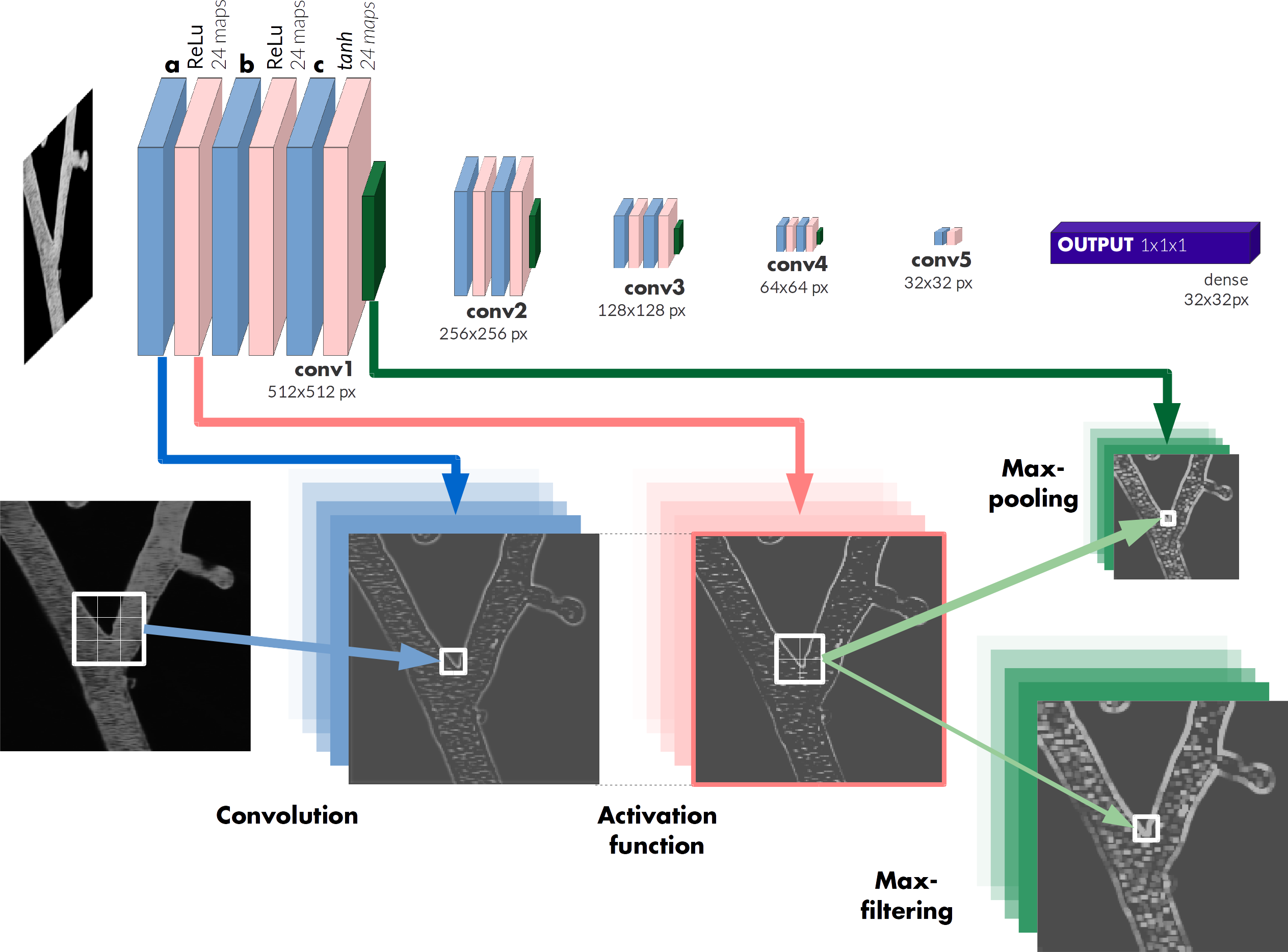}}

\caption{Example of a typical deep convolutional neural network (CNN)\textbf{
}using two-dimensional image as an example\textbf{. (top) }Deep neural
network consist of several subsequent layers (conv1, conv2, conv3,
conv4, conv5 in our example) of which each can contain stacked convolutional
layers (e.g. conv1a, conv1b, conv1c) that are followed by a non-linear
activation function which in our example are Rectified Linear Unit
(ReLU), and hyperbolic tangent (\emph{$tanh$}). The \emph{depth}
of the network is defined by the amount of layers, whereas the \emph{width
}of the network depend on the amount of feature maps generated on
each layer which in our case is 24 feature maps. The number of feature
maps correspond to the number of different learned convolutional kernels
on each convolutional layer, thus each conv1a, conv1b, conv1c have
24 different learned convolution kernel that try to represent the
training data. In our example the size of the convolution kernel is
3$\times3$ (see \textbf{bottom }the 3$\times3$ grid overlaid on
input image). Output of each layer is typically downsampled via max-pooling
operator that in our example takes the maximum value of 2$\times2$
window, thus the downsampling factor is 2 on each layer resulting
a total downsampling factor of 16 after 5 layers.\textbf{(bottom)
}The pipeline for one convolutional kernel (3$\times3$) is illustrated
for one feature map with edges enhanced which is then mapped with
$tanh$ activation function. The mapped feature map is then downsampled
using max-pooling operator (example in \textbf{top}), or alternatively
max-filtering can be applied as we will be using in this work that
does not change the image resolution allowing us to do dense image
segmentation without having to upsample the segmentation back to input
resolution. \textbf{}\label{fig:A-simple-convolutional}}
\end{figure*}

ConvNets are loosely inspired of biological networks (e.g. \cite{chen2014incremental})
allowing hierarchical feature learning starting from low-level features
such as edges into higher-level features such as faces for example.
ConvNets possess two key properties that make them useful in image
analysis: spatially shared weights and spatial pooling (\cite{pinheiro2013recurrent}).
This allows feature learning that is shift-invariant, i.e. filter
that is useful across the entire image as image statistics are stationary
\cite{simoncelli2001natural}. Typical convolutional networks are
composed of multiple stages (\figref{A-simple-convolutional}), and
the output of each stage is made of two or three dimensional arrays
depending on the training data, called feature maps. Each feature
map is the output of one convolutional filter (or pooling) applied
over the full image. This is typically followed by non-linear activation
function such as sigmoid, rectifying linear unit (ReLU) or hyperbolic
tangent (\emph{$tanh$}). 

After the final pooling layer of the network, there might be one or
more fully-connected (FC) layers that aim to perform high-level reasoning.
They take all neurons from the previous layer and connect them to
every single neuron of current layer (i.e. fully-connected). No spatial
information is preserved in typical fully-connected layer configurations.
In the end of the networks there is typically a terminal (``output'')
classification layer that based on the number of classes produces
real-valued or binary scalar for each image in image classification
dataset, or for the each pixel in each image image segmentation dataset.\textbf{
}The most typical output layer uses a \emph{softmax }regression that
generates probability distribution of the outputs \cite{gu2015recentadvances}.
The shortcoming of softmax is that does not capture model uncertainty
and often it is interpreted erroneously as model confidence \cite{gal2015dropout}.
If model uncertainty is needed, there have been effort to cast deep
learning models as Bayesian models\cite{gal2015dropout}. The networks
are typically regularized to mitigate over-fitting either using technique
called DropOut \cite{srivastava2014dropout} in which each neuron
has a probability of 0.5 to be reset with 0-value, typically only
used in last fully-connected layers. Alternatively one can regularize
the network by injecting noise for example just before the nonlinear
activation function \cite{poole2014analyzing}.

ConvNets are typically trained using stochastic gradient descent (SDG)
optimization method with \emph{mini-batches }so that the gradient
on each training iteration is computed using more than one training
example (i.e. patch of image/volume) resulting in smoother convergence,
and more efficient use of vectorization libraries, thus faster computation
times. ConvNets can be roughly divided to two basic types \cite{yuste2015fromthe}:
feedforward networks which are organized in layers with unidirectional
connections (e.g. the proposed approach here from Lee \emph{et al.}
\cite{lee2015recursive}), and recurrent network in which feedback
connectivity is dominant (e.g. used by Pinheiro \emph{et al.} \cite{pinheiro2013recurrent}
for semantic segmentation). Feedforward networks are typically used
for image classification and segmentation, whereas recurrent networks
are used for sequential data such as language, and sound processing.

Surprisingly even though the ConvNets have been highly successful,
the success of the ConvNets are not well understood even by the people
designing new algorithms and architectures (e.g. \cite{gu2015recentadvances}).
The ultimate goal of artificial intelligence (AI) including image
segmentation would be to build machines that understand the world
around us, i.e. disentangle the factors and causes it involves (\cite{bengio2012bettermixing}),
or in more practical terms, to have an image segmentation system that
would have an ``understanding'' of the vesselness. In our case eventually
exceeding the human expertise in determining which part of the image
is part of the vessel. This human-level concept learning was recently
demonstrated for written character recognition by Lake \emph{et al.}
\cite{lake2015humanlevel} from very limited training samples starting
from just one examples. For ``brute-force approaches'', there have
been ConvNets that have surpassed human-level performance on image
classification \cite{he2015delving,ioffe2015batchnormalization}.

We aim to improve the accuracy of the vessel segmentation for multiphoton
microscopy by training a deep learning framework based on convolutional
networks (ConvNets) in supervised manner with no free parameters for
the user to adjust. We have implemented our three-dimension vessel
segmentation using open-source CPU-accelerated ZNN framework \cite{lee2015recursive,zlateski2015imagesegmentation}
previously used for three-dimensional electron microscope segmentation.
Our main motivation for this proof-of-concept work is to inspire more
researchers to work on biomedical segmentation problems by providing
public available annotated dataset of two-photon fluorescence microscopy
vasculature stacks with the code needed to easily fine-tune the network
using your own training data and improve our model. 

Our work tries to integrate the fields of machine learning, biomedical
image analysis, and neuroscience\emph{ }and motivating applications.
Two-photon microscopy is capable of providing beautiful high-resolution
images of biological processes \emph{in vivo}. By creating an open
source, reproducible method of vascular segmentation, quantitative
results can be more readily attained and compared. We hope to decrease
the time overhead required for image processing by the average microscope
user and accelerate the educational translation of new information
to the scientific community.

\emph{}

\section{Related work}

\begin{figure*}
\textbf{\includegraphics[width=2\columnwidth]{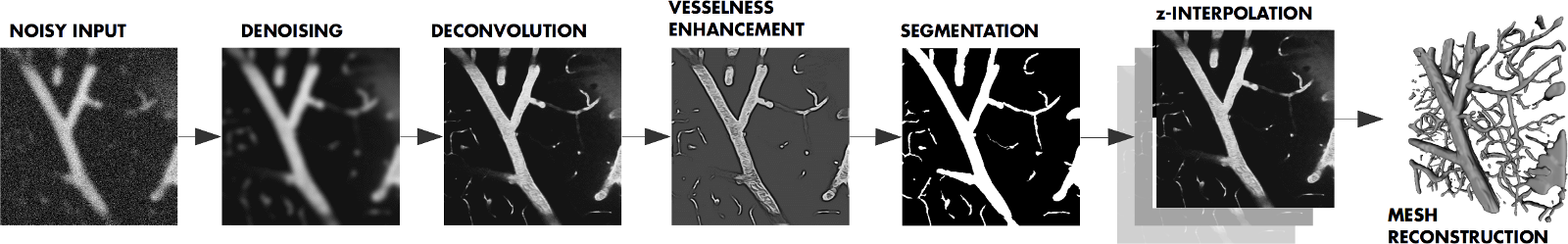}}\caption{Typical vessel segmentation pipeline as simplified schematic for a
single slice of a two-photon microscope image of mouse cortical vasculature.
The Poisson-corrupted image is denoised (e.g. BM4D \cite{maggioni2013nonlocal}),
and then deconvolved (e.g. blind Richardson-Lucy deconvolution \cite{mondal2014imagereconstruction}),
or this image restoration can be done jointly. This is followed by
a vesselness enhancement filter such as Frangi's filter \cite{frangi1998multiscale},
Optimal Oriented Flux (OOF) \cite{law2008threedimensional}, or some
more recent method (e.g by Moreno \emph{et al.} \cite{moreno2015gradientbased}).
This is followed by a segmentation algorithm (e.g. active contours
\cite{law2010anoriented}) that produces a binary mask (or real-valued
probability mask) that can be used to weigh the input. This weighed
image stack might be then interpolated in $z$-direction to obtain
an isotropic stack with equal-sized voxel sides for example using
a B-spline interpolation \cite{thevenaz2000interpolation}. If needed
for analysis or visualization, this interpolated stack can be reconstructed
into a three-dimensional mesh that is typically obtained via the Marching
Cubes algorithm variants \cite{levine2012meshprocessing2}.\label{fig:Typical-vessel-segmentation-pipeline}}
\end{figure*}

Typical simplified schematic of vasculature segmentation pipeline
used to process two-photon microscope stacks is shown in \figref{Typical-vessel-segmentation-pipeline}.
The image stacks suffer mainly from photon noise following a Poisson
distribution \cite{bertero2009imagedeblurring} (i.e. the noise intensity
depends on the underlying signal) with some Gaussian noise component
added, which can be denoised directly with methods developed for Poisson
noise (e.g. PURE-LET \cite{luisier2011imagedenoising}). Alternatively
the signal-dependency of the Poisson noise can be removed with a suitable
transform such as Anscombe transform \cite{makitalo2011optimal} that
allows one to use denoising methods developed for Gaussian noise (e.g.
BM3D/BM4D \cite{maggioni2013nonlocal,danielyan2014denoising}). Deconvolution
is not done as commonly for multiphoton microscopy as compared to
confocal microscopy \cite{mondal2014imagereconstruction}, but if
it is been done in can be done jointly with other image restoration
operations \cite{persch2013enhancing} or as its independent step
\cite{kim2015blinddepthvariant}. This part can be seen as the image
restoration part with an attempt to recover the ``original image''
as well as possible corrupted by the imaging process.

In some cases the restored image is further simplified using some
edge-aware smoothing operator such as anisotropic diffusion \cite{meijering2001evaluation,prasath2015multiscale},
or as done by Persch \emph{et al.} \cite{persch2013enhancing} who
jointly apply the anisotropic diffusion inpainting (operation that
attempts to replace lost or corrupted parts of the image data) with
deconvolution and interpolation. 

This step is followed by some ``vesselness filter'' or ``vesselness
enhancement'' filter that is designed to enhance tubular structures
such as vessels in the image. The best known filter of those is the
Frangi's filter \cite{frangi1998multiscale} that has become outdated
as it cannot properly handle crossings nor bifurcation methods, and
several filters \cite{law2008threedimensional,turetken2013detecting,smistad2013gpuaccelerated,hannink2014vesselness,moreno2015gradientbased}
have been proposed to correct the shortcomings of Frangi's filter
with none of them reaching a \emph{de facto }standard status. 

In our proposed deep learning-based network we are trying to replace
the vessel enhancement and segmentation steps, and keep still using
"traditional" filters with the image restoration part (see discussion
on how to get upgrade them as well in \ref{sub:Training-everywhere}).
There have been various "traditional" segmentation algorithms
for vessel segmentations (for reviews see \cite{kirbas2004areview,lesage2009areview}),
and only the most relevant ones are analyzed here below.

In the schematic (\figref{Typical-vessel-segmentation-pipeline})
$z$-interpolation is placed after the segmentation, but it might
have been placed as well before the segmentation algorithm \cite{lindvere2013cerebral,ukwatta20133dcarotid},
or jointly with other image restoration operators \cite{persch2013enhancing}.
The exact placing of the interpolation depends on the computation
before and after it, but in our case we placed in the end to emphasize
the gains of $z$-direction interpolation to mesh reconstruction as
all our stacks used in this work are anisotropic (see \tabref{Dataset-used-in-Study}).
Reconstructing meshes from non-interpolated anisotropic stacks with
traditional Marching Cubes algorithm \cite{levine2012meshprocessing2}
typically leads to ``staircasing effect'' of the mesh while interpolation
gives smoother reconstruction. Advanced mesh reconstruction methods
are beyond the scope of this algorithm, but there have been efforts
to improve biomedical mesh reconstruction \cite{moriconi2015highquality,saha2015digital}
mitigating the problems of triangulation based such as Marching Cubes.
With the reconstructed vasculature mesh, it is then possible to for
example do morphological analysis \cite{meyer2008altered}, calculate
hemodynamic parameters \cite{keshmiri2015vascular}, or analyze the
functional diameter changes in response to external stimulus \cite{lindvere2013cerebral}.

To the knowledge of the authors, deep learning frameworks including
ConvNets have not yet been applied to segmentation of three-dimensional
volumetric vasculature images. Despite the limited use of machine
learning techniques in VESSEL12 challenge for lung vessels \cite{rudyanto2014comparing},
there have been some work using machine learning techniques for vessel
segmentation. Sironi \emph{et al.} \cite{sironi2015learning} for
example used an unsupervised dictionary learning \cite{kreutz-delgado2003dictionary}
approach that learned optimal separable convolutional filter banks
for 2D vasculature segmentation (DRIVE dataset \cite{staal2004ridgebased}),
and for 3D olfactory projection fibers (DIADEM challenge \cite{brown2011thediadem}).
The filter banks were then used with the popular Random Forests classifier
\cite{breiman2001randomforests} continuing the previous work from
the same lab \cite{gonzalez2009learning,rigamonti2012accurate}. The
authors used their separable filter banks with ConvNets for image
classification task but did not discuss about the possibility of using
ConvNets with the image segmentation task. Very recently Maji et al.
\cite{maji2016ensemble} applied ConvNets for the two-dimensional
vasculature DRIVE database with promising performance. 

Santamaria-Pang \emph{et al.} \cite{santamaria-pang2007automatic}
similarly used a dictionary learning approach to learn linear filters
for detection of tubular-like structures from multiphoton microscopy
stacks. The learned filters were fed to a Support Vector Machine (SVM,
\cite{suykens1999leastsquares}) which was shown to provide a better
segmentation accuracy compared to the vesselness filter introduced
by Sato \emph{et al.} \cite{sato19973dmultiscale}. Recently, Schneider
\emph{et al.} \cite{schneider2015joint3d} used Random Forests for
classification with multivariate Hough forests to infer probabilistic
votes about the vessel center, jointly segmenting vasculature and
extracting vessel centerline. The features were learned using steerable
filter templates (\cite{jacob2004designof}) at multiple scales instead
of the dictionary learning approach. They showed that their learning-based
approach outperformed both Oriented Optimal Flow (OOF, \cite{law2008threedimensional})
and Frangi's filter \cite{frangi1998multiscale} for vessel segmentation.

Sironi \emph{et al.} \cite{sironi2015projection} take a different
approach in their paper inspired by recent work on structured learning-based
edge detectors (\cite{dollar2015fastedge}). They combine structured
learning with nearest neighbor-based output refinement step designed
for situations where edges or thin objects are hard to detect explicitly
by the neural network (\cite{ganin2014ntextasciicircum4fields}).
 They were able to reduce spatial discontinuities, isolated erroneous
responses and topological errors of initial score maps from outputs
of other algorithms, and when directly trained to segment two-dimensional
blood vessels (DRIVE dataset \cite{staal2004ridgebased}).

There is relatively more work devoted on natural image processing
compared to biomedical image analysis. In natural image processing
literature, the corresponding application to our biomedical image
segmentation is semantic segmentation \cite{long2014fullyconvolutional,papandreou2015weaklyand,chen2015attention,chen2015semantic},
also referred as scene parsing \cite{pinheiro2013recurrent} or scene
labeling \cite{farabet2013learning}. Semantic segmentation with natural
images tries to answer to the question ``What is where in your image?''
for example segmenting the ``driver view'' in autonomous driving
to road, lanes and other vehicles \cite{kendall2015bayesian}. In
typical semantic segmentation tasks there are a lot more possible
labels than in our two-label segmentation of vessels and non-vessel
voxels, further complicating the segmentation.

Most existing biomedical segmentation pipelines start with slice-by-slice
two-dimensional processing for volumetric stacks, and only later transition
to three-dimensional processing due to high computational cost of
fully three-dimensional pipelines \cite{liu2014amodular,takemura2013avisual}.
ConvNets with 3D filters had been used for example with block face
EM images before \cite{helmstaedter2013connectomic}, most of the
3D filter use being employed in video processing \cite{ji20133dconvolutional,tran2015learning,yao2015describing}
where the 2D image with the time can be viewed as an anisotropic 3D
image. Due to ever-increasing computational performance in local GPU
clusters, and cloud-based services such as Amazon AWS, IBM Softlayer,
Microsoft Azure and Google Cloud Platform we expect to see more purely
three-dimensional approaches such as the one proposed by Kamnitsas
\emph{et al. }\cite{kamnitsas2016efficient} for\emph{ }brain lesion
segmentation from MRI images.

Deep learning based approaches have been extensively used for volumetric
electron microscopy (EM) segmentation \cite{huang2013deepand,maitin-shepard2015combinatorial,wu2015aniterative,lee2015recursive,ronneberger2015unetconvolutional}.
Other biomedical image segmentation tasks with deep learning frameworks
include for example brain segmentation \cite{havaei2015braintumor,lyksborg2015anensemble,kamnitsas2015multiscale,stollenga2015parallel},
prediction of Alzheimer's disease from magnetic resonance imaging
(MRI) scans \cite{payan2015predicting}, microscopic cell segmentation
\cite{kraus2015classifying}, glaucoma detection \cite{chen2015automatic},
computational mammography \cite{dubrovina2015computational}, pancreas
segmentation \cite{dubrovina2015computational}, bi-ventrical volume
estimation \cite{zhen2015multiscale}, and carotid artery bifurcation
detection \cite{zheng20153ddeep}

The use of deep learning neural networks is not limited to image analysis,
and it can employed in various fields that can benefit from data-driven
analysis in exploratory or predictive fashion. In neuroscience, in
general the datasets are getting increasingly larger and more complex
requiring more sophisticated data analysis tools \cite{rubinov2015neuralnetworks}.
There have been systems capable of constructing theories automatically
in data-driven fashion \cite{ghahramani2015probabilistic}. Artificial
neural networks lend themselves well  for modeling complex brain
function that emerge from activation of ensembles of neurons in which
the studying of single neuron at a time is not sufficient \cite{rubinov2015neuralnetworks}. 

For example, the circuit architecture of the mammalian hippocampus
have been modeled to consist of series of sequential feedforward and
recurrent neural networks \cite{rolls1998neuralnetworks}. Harvey
\emph{et al. }\cite{harvey2012choicespecific} used two-photon imaging
to measure the calcium activity of mouse making behavioral choices
in virtual maze. The temporal trajectory of neuron populations was
shown to be predictive of the behavioral choice, thus being suitable
for the use of recurrent neural networks to model the behavior. In
addition to basic neuroscience, deep learning ``expert systems''
have been extended to clinical settings \cite{waljee2010machine}
for example for predicting clinical outcomes of radiation therapy
\cite{kang2015machine}, electroencephalographic (EEG) recording
analysis \cite{stober2015deepfeature}, and future disease diagnosis
and medicine prescription in routine clinical practice \cite{choi2015doctorai}.

\section{Methods}

\subsection{Dataset }

\begin{figure}
\centerline{\includegraphics[width=0.9\columnwidth]{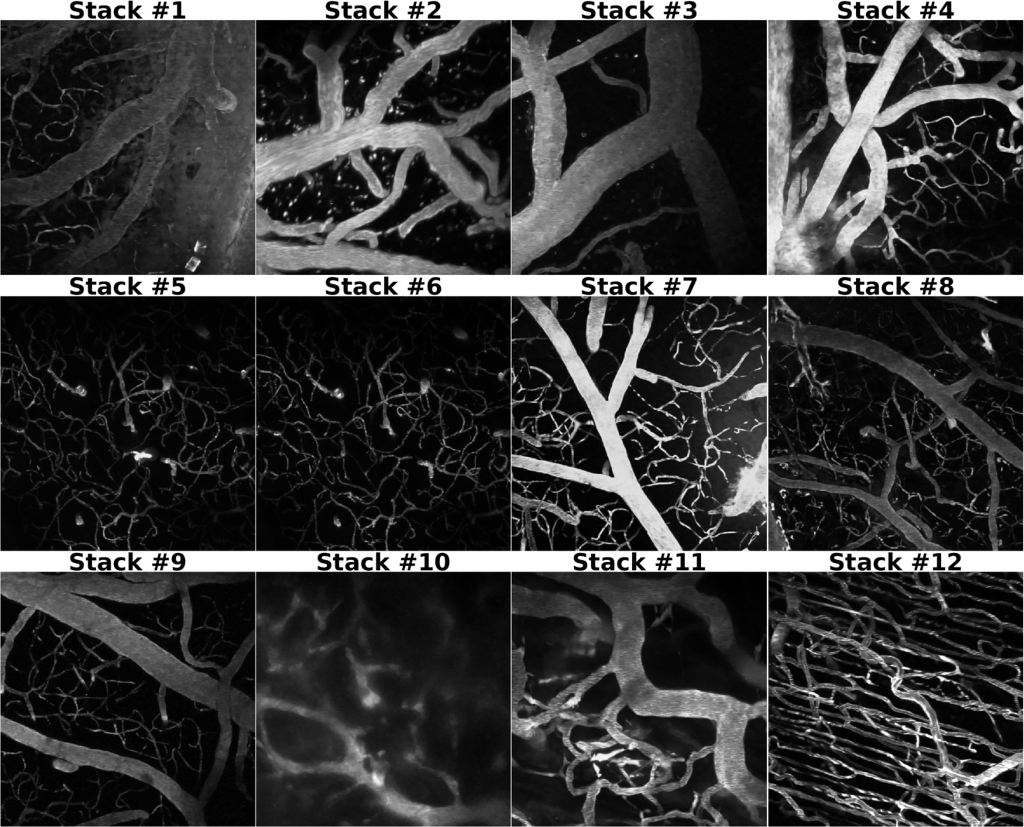}}

\caption{Training dataset visualized as maximum-intensity projections (MIP).
The Stack \#5 and Stack \#6 are acquired from same experimental session
on different time points with Stack \#5 showing fluorescent dye leakage
due to focused ultrasound stimulation. Stack \#10 turned out to be
too hard for our network to segment properly, and the network would
need more similar training data to handle inferior image quality as
well.\textbf{\label{fig:Training-dataset-visualized}}}
\end{figure}

The vessel dataset described here were acquired from mouse cortex,
and from GFP-labelled human squamous cell carcinoma tumors, xenografted
onto the dorsal skin of mice with implanted dorsal window chambers
(FaDu-GFP, AntiCancer Inc.), tumors summarized in \tabref{Dataset-used-in-Study}
(see the maximum-intensity projections of stacks in \ref{fig:Training-dataset-visualized}).
Fluorescent dextran (70 kDa Texas Red, dissolved in PBS, Invitrogen)
was used to visualize the vasculature in mouse cortex by \cite{burgess2014analysis},
and fluorescent dextran (2MDa FITC, dissolved in PBS, Invitrogen)
to label the tumor vasculature. Imaging was performed using the FV1000
MPE two-photon laser scanning microscope (Olympus) with tunable mode-locked
Ti:Sapphire laser using several excitation wavelengths and water-immersion
objective lenses\emph{.}

The auxiliary Matlab code for our implementation of ZNN is provided in \href{https://github.com/petteriTeikari/vesselNN}{https://github.com/petteriTeikari/vesselNN},
with the annotated dataset available from \href{https://github.com/petteriTeikari/vesselNN_dataset}{https://github.com/petteriTeikari/vesselNN\_dataset}.

\begin{table*}
\caption{Dataset used in the study (check resolution from .oib files, and re-denoise
the image at some point with correct metadata as ImageJ lost it).
Additional possible parameters: \emph{depth}, \emph{FOV}, and\emph{
dye, excitation wavelength, percentage of vessel labels (see \cite{ciresan2012deepneural}).
}\label{tab:Dataset-used-in-Study}}

\scriptsize{%
\begin{tabular*}{2\columnwidth}{@{\extracolsep{\fill}}c|cccccc}
\# & \textbf{Resolution ($\mu$m\textsuperscript{3})} & \textbf{Dimension (voxel\textsuperscript{3})} & \textbf{\# samples} & \textbf{\% of vessel labels} & \textbf{Source} & \textbf{Usage}\tabularnewline
\hline 
1 & 0.994$\times$0.994$\times$5 & 512$\times$512$\times$15 & 3.75M & 12.4\% & Mouse cortex & Train\tabularnewline
2 & 1.59$\times$1.59$\times$5 & 320$\times$320$\times$26 & 2.54M & 29.8\% & Mouse cortex & Train\tabularnewline
3 & 0.994$\times$0.994$\times$5 & 512$\times$512$\times$10 & 2.5M & 42.1\% & Mouse cortex & Train\tabularnewline
4 & 0.994$\times$0.994$\times$5 & 512$\times$512$\times$15 & 3.75M & 36.1\% & Mouse cortex & Train\tabularnewline
5 & 0.994$\times$0.994$\times$5 & 512$\times$512$\times$25 & 6.25M & 3.2\% & Mouse cortex & Train\tabularnewline
6 & 0.994$\times$0.994$\times$5 & 512$\times$512$\times$25 & 6.25M & 3.7\% & Mouse cortex & Test\tabularnewline
7 & 0.994$\times$0.994$\times$5 & 512$\times$512$\times$23 & 5.75M & 9.5\% & Mouse cortex & Test\tabularnewline
8 & 0.994$\times$0.994$\times$5 & 512$\times$512$\times$25 & 6.25M & 9.0\% & Mouse cortex & Train\tabularnewline
9 & 2.485$\times$2.485$\times$5 & 512$\times$512$\times$14 & 3.5M & 34.0\% & Mouse cortex & Train\tabularnewline
10 & 0.621$\times$0.621$\times$5 & 512$\times$512$\times$15 & 3.75M & 10.5\% & Tumor & Train\tabularnewline
11 & 0.621$\times$0.621$\times$5 & 512$\times$512$\times$21 & 5.25M & 24.1\% & Tumor & Train\tabularnewline
12 & 0.621$\times$0.621$\times$5 & 512$\times$512$\times$27 & 6.75M & 14.2\% & Tumor & Train\tabularnewline
\end{tabular*}}
\end{table*}

\subsubsection{Data import}

\label{sub:Data-import}We used the Java-based Bio-Formats library
(OME - The Open Microscopy Environment, \href{https://www.openmicroscopy.org/}{https://www.openmicroscopy.org/},
\cite{linkert2010metadata,moore2015omeroand}) with Matlab\cite{li2015metadata}
to open the OIB files from Olympus FluoView 2-photon microsopy setup.
We selected representative substacks from each original stack to reduce
the time needed for manual annotation by us researchers. The substacks
were converted to 16-bit OME-TIFF image files containing all the original
metadata.

\subsubsection{Data annotation}

The ground truth for the vessels were manually annotated slice-by-slice
using custom-written Matlab code to produce a ``seed binary'' image
containing the strongest edges which then had to be refined manually
using the pencil tool of GIMP (\href{http://www.gimp.org}{http://www.gimp.org}).
We used more conservative criteria for labeling vasculature than the
traditional ``50\% of the voxel'' to account the partial volume
effect \cite{taha2015metrics}, and we tried to include all the vessel-like
structures to the label mask.

\subsubsection{Denoising (Image Restoration)}

After converting the substacks to OME-TIFF files, we denoised the
microscopy stacks using the state-of-the art denoising algorithm BM4D
(\cite{maggioni2013nonlocal}) implemented in Matlab. BM4D is a volumetric
extension of the commonly used BM3D denoising algorithm \cite{dabov2007imagedenoising}
for 2D images, which was for example used to denoise two-photon microscope
images by Danielyan \emph{et al.} \cite{danielyan2014denoising}.
They also demonstrated that the two-photon microscopy noise can be
modeled well using the models developed for digital cameras. BM3D/BM4D
were designed for denoising images degraded by Gaussian noise, thus
we applied first Anscombe transform to reduce the signal-dependency
of the noise as done with BM4D for denoising of magnetic resonance
imaging (MRI) images \cite{makitalo2011optimal}. After the BM4D denoising,
an inverse Anscombe transform was applied to convert the stacks back
to original intensity domain.

Two of the stacks (\texttt{burgess2014 bbbDisruption}, and \texttt{burgess2014 noisySparseVessels})
were degraded by horizontal periodic ``banding'' caused by improperly
balanced microscope stage, and the degradation was mitigated using
spatial notch filters in frequency domain applying fast Fourier Transform
(FFT) in Matlab. Noise components were manually identified and then
removed before denoising those images. We did not apply any blind
deconvolution (e.g. \cite{dupe2009aproximal}) for our microscope
stacks to improve the image quality. There was no significant spectral
crosstalk in any of the stacks, thus no spectral unmixing or blind
image separation (e.g. \cite{dao2014useof}) was done for the image
stacks. Likewise, no motion compensation algorithms (e.g. \cite{soulet2013automated}) was needed for the dataset.

\subsubsection{Error metrics}

\label{sub:Error-metrics}To analyze the segmentation quality of our
proposed architecture we used Average Hausdorff Distance (AVD) as
the error metric. The AVD between the ground truth and output of the
proposed architecture was computed using the \texttt{EvaluateSegmentation}
package (\href{http://github.com/codalab/EvaluateSegmentation}{http://github.com/codalab/EvaluateSegmentation})
published by Taha \emph{et al. }\cite{taha2015metrics}. AVD was chosen
as the metric as it is well suited for evaluating complex boundary
delimitation. Disadvantage of the AVD is that it is based on calculating
the distances between all pairs of voxels, making it computationally
intensive and not feasible to be integrated to network training for
example.

\textbf{}

\subsection{Deep learning network}

We trained our 3D vessel segmentation deep ConvNet using ZNN framework
\cite{zlateski2015znn}, that uses multicore CPU parallelism for speed
instead of typical GPU-accelerated frameworks such as Theano for example
\cite{thetheanodevelopmentteam2016theanoa}. At the time of our training,
there were not many frameworks available that would take the 3D context
into account. Commonly used library Caffe \cite{jia2014caffeconvolutional}
had only 2D networks available, while DeepLab built on top of Caffe
would have had GPU-accelerated 3D networks implemented. Our approach
for vessel segmentation is inspired by the success of ZNN in segmenting
three-dimensional electron microscope (EM) image stacks \cite{lee2015recursive},
and we chose to start with the networks described for EM segmentation.

\subsubsection{Training with ZNN}

ZNN produces a dense output with pixel-by-pixel segmentation maps
in contrast to image-level labels in object recognition. ConvNets
have excelled in object recognition which typically only require single
output value for an entire input image \emph{{[}i.e. is there a dog
in the image? yes (1), or no (0){]}}. ZNN employs max-filtering which
slides a window across image and applies the maximum operation to
that window retaining original image resolution. Traditionally, in
semantic segmentation (\cite{long2014fullyconvolutional}) and biomedical
image segmentation (\cite{ronneberger2015unetconvolutional}) pipelines,
max-pooling is used instead of max-filtering which reduces the dimensions
of the output map requiring either post-processing using for example
some graphical model (\cite{chen2015semantic}), or upsampling back
to the original resolution (\cite{ronneberger2015unetconvolutional})\emph{.
}The max-filtering employed by ZNN can be thought as the dense variant
of max-pooling filter as it keeps image dimensions intact while making
all filtering operation sparse both via convolution and max-filtering.
This approach is also called “skip-kernels”\emph{ }(\cite{sermanet2013overfeat})
or “filter rarefaction” (\cite{long2014fullyconvolutional}), and
is equivalent in its results to “max-fragmentation-pooling” (\cite{giusti2013fastimage,masci2013afast}).
In practice with ZNN we can control the sparseness of filters independent
of max-filtering.

\subsubsection{Network architecture}

We adopted the recursive architecture from Lee \emph{et al.} \cite{lee2015recursive}
used to segment electron microscopy (EM) stacks. 

\begin{figure}
\includegraphics[width=1\columnwidth]{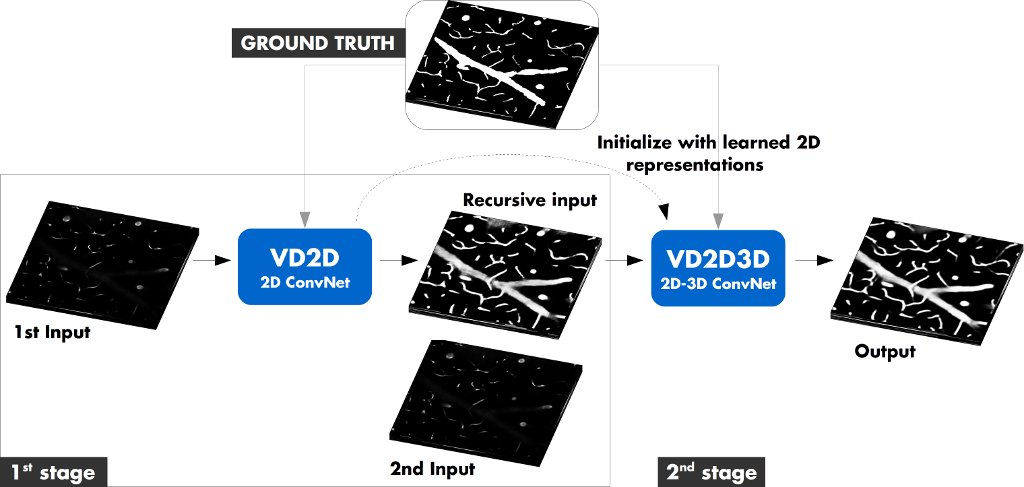}

\caption{An overview of our proposed framework (left) and model architectures
(right,). The number of trainable parameters in each model is 230K
(VD2D), 310K (VD2D3D).(\cite{lee2015recursive}).\label{fig:An-overview-of-ZNN-architecture}}
\end{figure}

\begin{figure*}
\centerline{\includegraphics[width=1.8\columnwidth]{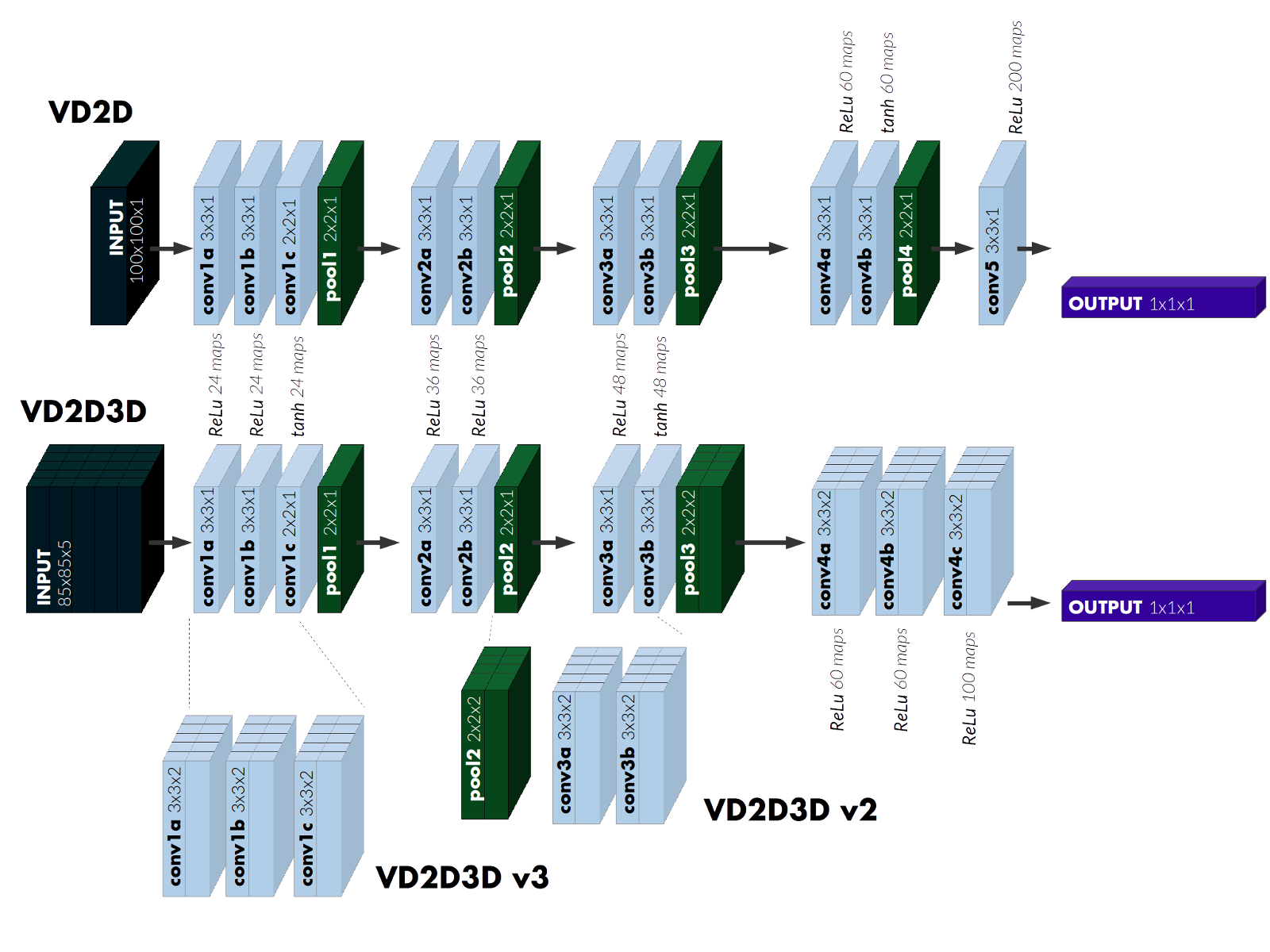}}

\caption{Network architectures of the different used models: VD2D, VD2D3D,
and the extensions of the latter VD2D3D\_v2 and VD2D3D\_v3 which had
some of the two-dimensional convolution filters converted to three-dimensional
filters. \label{fig:Network-architectures-of-ZNN}}
\end{figure*}

\paragraph{VD2D}

The chosen recursive architecture first involved a two-dimensional
VD2D (``Very Deep 2D'') ``pre-training'' stage that is shown in
\figref{Network-architectures-of-ZNN} and in \figref{An-overview-of-ZNN-architecture}.
All convolutions filters have sizes of 3$\times$3$\times$1, except
that \texttt{Conv1c} uses a 2$\times$2$\times$1 filter to make
the ``receptive field'' for a single output pixel to have an odd-numbered
size, and thus centerable around the output pixel\emph{. }Some convolution
layers are employing hyperbolic tangent ($tanh$) nonlinearities rather
than traditionally used rectifying linear units (ReLUs) as the authors
argued that this might suppress variations in the feature maps due
to image quality variations. This was left however untested in their
original paper.

\paragraph{VD2D3D}

The two-dimensional convolutional layers of the following second stage
named VD2D3D (``Very Deep 2D-3D'', see \figref{Network-architectures-of-ZNN}
and \figref{An-overview-of-ZNN-architecture}) are initialized with
the trained weights of the VD2D without enforcing weight sharing as
done by some recurrent ConvNets (\cite{pinheiro2013recurrent}). The
main idea behind having initial 2D layers in the VD2D3D is to make
the network faster to run and train, while the 3D filters in the layers
enable the network to use 3D context in vessel segmentation providing
more accurate predictions. 

In theory the accuracy could be further improved by transforming all
the layers to 3D but this would in practice come with increased computational
cost and memory requirements. The VD2D3D could be used directly for
the denoised input images without the initial VD2D training, but Lee
\emph{et al. }\cite{lee2015recursive} showed that providing the output
of VD2D recursively as the input to the VD2D3D produced a significant
improvement in performance. The layers \texttt{Conv1a}, \texttt{Conv1b},
and \texttt{Conv1c} are used to process the recursive inputs along
with the denoised input images, which then are combined together after
\texttt{Conv1c}. This parallel processing stream should allow more
complex, highly nonlinear interaction between low-level features and
contextual information in the recursive input. The increase of trainable
parameters due to switch from 2D filters to 3D filters were compensated
by trimming the size of later layer feature map from 200 (\texttt{Conv5}
of VD2D) to 100 (\texttt{Conv4c} of VD2D3D)

\paragraph{VD2D3D\_v2}

We changed the last two-dimensional layer (\texttt{Conv3} into three-dimensional
layer (see VD2D3D\_v2 in \figref{Network-architectures-of-ZNN}) keeping
the VD2D3D otherwise the same.

\paragraph{VD2D3D\_v3}

.We wanted to see what would be the effect of changing the first layer
into three-dimensional. This in practice would correspond to the low-level
features and should improve the detection of three-dimensional structures
rather over two-dimensional filters that could confuse ``feature-like''
two-dimensional noise to ``real'' three-dimensional vasculature.

\subsubsection{Training procedure}

\label{sub:Training-procedure}The network training procedure was
similar to the one described by Lee \emph{et al.} \cite{lee2015recursive}.
We trained our network using backpropagation with the cross-entropy
loss function. The VD2D was first trained for 60K updates using 100$\times$100$\times$1
output patches. The initial learning rate was set to 0.01, the momentum
of 0.9, and an annealing factor of 0.999 which was applied every 6
updates giving us a learning rate of 0.000000452 at the end of VD2D
training. Each update took around 2.9 seconds in our Intel Dual Intel
Xeon E5650\textbf{ }Quad CPU\textbf{ }(16 hyperthreads, 24 GB RAM)
workstation on Ubuntu 14.04, with all the 16 threads in use giving
us a total of 2 days for the VD2D training. After completing VD2D
training, we continued with the training of VD2D3D for 90,000 updates
as in the original paper by Lee \emph{et al. }\cite{lee2015recursive}
with an initial learning rate of 0.01, the momentum of 0.9 and with
the same annealing factor of 0.999 which was applied on every update
for 15K updates, after which the learning rate was set 0.0001 with
the same annealing factor that was this time applied on every 10th
update. Each update took around 23 seconds, giving us a total of 24
days for the training of VD2D3D with the same 90K updates.

For the modified architectures with extended 3D support (v2 and v3)
higher memory were required, and fully 3D pipeline was not possible
with the current implementation of ZNN with just 24 GB of RAM. Each
update with v2 took around 27.2 seconds (90,000 updates took slightly
over 28 days), and with v3 each update took around 24.4 /seconds (90,000
updates took slightly over 25 hours).

Like Lee \emph{et al.} \cite{lee2015recursive}, we rebalanced the
classes (vessels/non-vessels) by differentially weighing the per-pixel
loss to deal with the imbalance between vessels and non-vessel pixels
which was however lower than the imbalance seen in electron microscope
images between boundary and non-boundary pixels. 

We also augmented the data by randomly rotating and flipping 2D image
patches as implemented in ZNN. Additionally we could have introduced
photometric distortions (\cite{howard2013someimprovements}) to further 
counteract the possible overfitting due to limited training data, but
they were seen unnecessary at the time of the training.

We also used dropout (\cite{srivastava2014dropout}) to further avoid
overfitting that was implemented in ZNN. Dropout was applied to the
\texttt{Conv4c} layer with a probability of 0.5 to be reset with
a 0-valued activation.

\subsection{Sharing}

Our proposed segmentation pipeline is based on the ZNN framework that
is freely available online at \href{https://github.com/seung-lab/znn-release}{https://github.com/seung-lab/znn-release}
by the original authors \cite{lee2015recursive,zlateski2015znn}.
We have develop some helper function for that using Matlab, and all
those files are available from our Github repository at \href{https://github.com/petteriTeikari/vesselNN}{https://github.com/petteriTeikari/vesselNN}.
In the spirit of reproducible research \cite{vandewalle2009reproducible,dechaumont2012icyan,kenall2015betterreporting,leek2015opinion}
we release also our annotated dataset for other research teams to
be used. The dataset is available from \href{https://github.com/petteriTeikari/vesselNN}{https://github.com/petteriTeikari/vesselNN}.

\section{Results}

\label{sec:Results}

See the summary of results of the training in \tabref{Summary-of-the-results}
which basically shows that VD2D3D is better than VD2D as expected,
and that stack 10 ruins the statistics as it was not segmented that
well. Otherwise the Average Hausdorff Distance might be a a bit abstract,
but smaller distance the better, and it was recommended for complex
boundaries such as vessels and neurons in the review by Taha and Hanbury
\cite{taha2015metrics}.

The more detailed results of VD2D and VD2D3D architecture with thresholding
and dense CRF post-processing can be seen in \tabref{Results-of-VD2D3D},
quantified using Hausdorff average distance (AVD). The difference
in performance between different variants of the VD2D3D and VD2D is
shown in \tabref{Summary-of-the-Variants}, quantified using the same
AVD metric. Comparison of different metrics for the baseline VD2D3D
is shown in \tabref{Results-of-VD2D3D} to provide better interpretability
compared to other studies as AVD is not the most typically used metric.
Rand Index and Area Under the Curve (AUC) was chosen as metrics as
they are typically used as error metrics in medical segmentation studies
\cite{taha2015metrics}. . Mutual information quantifies recall (i.e.
the segmentation should have all the regions marked in the ground
truth, while not penalizing the added regions too much) on cost of
precision. Hausdorff distance and Mahalanobis distance are spatial
distance based metrics closely related to our method of choice Average
Hausdorff Distance (AVD) that is basically a more robust version of
Hausdorff distance handling outliers better. Mahalanobis distance
would be preferred in segmentation where general shape and alignment
are important.

\begin{table*}[t]
\caption{Summary of the results using Hausdorff average distance (AVD)
as the measure of segmentation quality. Thresholding is considered the worst-case scenario
and DenseCRF inference more advanced version for binary segmentation.\label{tab:Summary-of-the-results}}

{\scriptsize{}}%
\begin{tabular*}{2\columnwidth}{@{\extracolsep{\fill}}>{\centering}b{0.15\columnwidth}>{\centering}b{0.2\columnwidth}|>{\centering}b{0.05\columnwidth}>{\centering}b{0.05\columnwidth}
>{\centering}b{0.05\columnwidth}>{\centering}b{0.05\columnwidth}
>{\centering}b{0.05\columnwidth}>{\centering}b{0.05\columnwidth}
>{\centering}b{0.05\columnwidth}>{\centering}b{0.05\columnwidth}
>{\centering}b{0.05\columnwidth}>{\centering}b{0.05\columnwidth}
>{\centering}b{0.05\columnwidth}>{\centering}b{0.05\columnwidth}|>{\centering}b{0.05\columnwidth}>{\centering}b{0.05\columnwidth}}
\textbf{\scriptsize{}Network } & \multicolumn{1}{>{\centering}b{0.2\columnwidth}}{\textbf{\scriptsize{}Post-processing }} & \textbf{\scriptsize{}1 } & \textbf{\scriptsize{}2 } & \textbf{\scriptsize{}3 } & \textbf{\scriptsize{}4 } & \textbf{\scriptsize{}5 } & \textbf{\scriptsize{}6 } & \textbf{\scriptsize{}7 } & \textbf{\scriptsize{}8 } & \textbf{\scriptsize{}9 } & \textbf{\scriptsize{}10 } & \textbf{\scriptsize{}11 } & \multicolumn{1}{>{\centering}b{0.05\columnwidth}}{\textbf{\scriptsize{}12 }} & \textbf{\scriptsize{}Mean } & \textbf{\scriptsize{}SD }\tabularnewline
\hline 
{\scriptsize{}VD2D3D } & {\scriptsize{}DenseCRF 2D } & {\scriptsize{}1.44 } & {\scriptsize{}0.19 } & {\scriptsize{}0.23 } & {\scriptsize{}0.29 } & {\scriptsize{}0.67 } & {\scriptsize{}0.67 } & {\scriptsize{}0.48 } & {\scriptsize{}0.49 } & {\scriptsize{}0.18 } & {\scriptsize{}0.98 } & {\scriptsize{}0.98 } & {\scriptsize{}0.31 } & {\scriptsize{}0.57 } & {\scriptsize{}0.38 }\tabularnewline
 & {\scriptsize{}Thresholding } & {\scriptsize{}2.36 } & {\scriptsize{}0.46 } & {\scriptsize{}0.49 } & {\scriptsize{}0.35 } & {\scriptsize{}1.03 } & {\scriptsize{}1.05 } & {\scriptsize{}1.19 } & {\scriptsize{}1.18 } & {\scriptsize{}0.35 } & {\scriptsize{}1.66 } & {\scriptsize{}1.79 } & {\scriptsize{}0.62 } & {\scriptsize{}1.04 } & {\scriptsize{}0.61 }\tabularnewline
{\scriptsize{}VD2D} & {\scriptsize{}DenseCRF 2D } & {\scriptsize{}1.75 } & {\scriptsize{}0.20 } & {\scriptsize{}0.25 } & {\scriptsize{}0.25 } & {\scriptsize{}0.78 } & {\scriptsize{}0.83 } & {\scriptsize{}0.87 } & {\scriptsize{}0.63 } & {\scriptsize{}0.20 } & {\scriptsize{}1.08 } & {\scriptsize{}1.11 } & {\scriptsize{}0.34 } & {\scriptsize{}0.69 } & {\scriptsize{}0.46 }\tabularnewline
 & {\scriptsize{}Thresholding } & {\scriptsize{}2.58 } & {\scriptsize{}0.53 } & {\scriptsize{}1.30 } & {\scriptsize{}0.43 } & {\scriptsize{}1.32 } & {\scriptsize{}1.31 } & {\scriptsize{}1.44 } & {\scriptsize{}1.41 } & {\scriptsize{}0.47 } & {\scriptsize{}1.80 } & {\scriptsize{}1.98 } & {\scriptsize{}0.73 } & {\scriptsize{}1.28 } & {\scriptsize{}0.63 }\tabularnewline
\end{tabular*}
\end{table*}

\begin{table*}[t]
\caption{Summary of the results between different architecture variants using
Hausdorff average distance (AVD) as the measure of segmentation quality.
The best measure (the lowest value) for each individual stack and
for statistical value is shown in bold.\label{tab:Summary-of-the-Variants}}

{\scriptsize{}}%
\begin{tabular*}{2\columnwidth}{@{\extracolsep{\fill}}>{\centering}b{0.15\columnwidth}>{\centering}b{0.2\columnwidth}|>{\centering}b{0.05\columnwidth}>{\centering}b{0.05\columnwidth}
>{\centering}b{0.05\columnwidth}>{\centering}b{0.05\columnwidth}
>{\centering}b{0.05\columnwidth}>{\centering}b{0.05\columnwidth}
>{\centering}b{0.05\columnwidth}>{\centering}b{0.05\columnwidth}
>{\centering}b{0.05\columnwidth}>{\centering}b{0.05\columnwidth}
>{\centering}b{0.05\columnwidth}>{\centering}b{0.05\columnwidth}|>{\centering}b{0.05\columnwidth}>{\centering}b{0.05\columnwidth}}
\textbf{\scriptsize{}Network } & \multicolumn{1}{>{\centering}b{0.2\columnwidth}}{\textbf{\scriptsize{}Post-processing }} & \textbf{\scriptsize{}1 } & \textbf{\scriptsize{}2 } & \textbf{\scriptsize{}3 } & \textbf{\scriptsize{}4 } & \textbf{\scriptsize{}5 } & \textbf{\scriptsize{}6 } & \textbf{\scriptsize{}7 } & \textbf{\scriptsize{}8 } & \textbf{\scriptsize{}9 } & \textbf{\scriptsize{}10 } & \textbf{\scriptsize{}11 } & \multicolumn{1}{>{\centering}b{0.05\columnwidth}}{\textbf{\scriptsize{}12 }} & \textbf{\scriptsize{}Mean } & \textbf{\scriptsize{}SD }\tabularnewline
\hline 
{\scriptsize{}VD2D} & {\scriptsize{}DenseCRF 2D } & {\scriptsize{}1.75 } & {\scriptsize{}0.20 } & {\scriptsize{}0.25 } & {\scriptsize{}0.25 } & {\scriptsize{}0.78 } & {\scriptsize{}0.83 } & {\scriptsize{}0.87 } & {\scriptsize{}0.63 } & {\scriptsize{}0.20 } & {\scriptsize{}1.08 } & {\scriptsize{}1.11 } & {\scriptsize{}0.34 } & {\scriptsize{}0.69 } & {\scriptsize{}0.46 }\tabularnewline
{\scriptsize{}VD2D3D } & {\scriptsize{}DenseCRF 2D } & {\scriptsize{}1.44 } & {\scriptsize{}0.19 } & {\scriptsize{}0.23 } & {\scriptsize{}0.29 } & \textbf{\scriptsize{}0.67 } & {\scriptsize{}0.67 } & {\scriptsize{}0.48 } & {\scriptsize{}0.49 } & \textbf{\scriptsize{}0.18}{\scriptsize{} } & {\scriptsize{}0.98 } & {\scriptsize{}0.98 } & {\scriptsize{}0.31 } & {\scriptsize{}0.57 } & {\scriptsize{}0.38 }\tabularnewline
{\scriptsize{}VD2D3D\_v2} & {\scriptsize{}DenseCRF 2D } & \textbf{\scriptsize{}1.17 } & {\scriptsize{}0.20 } & {\scriptsize{}0.24 } & {\scriptsize{}0.30 } & {\scriptsize{}0.70 } & \textbf{\scriptsize{}0.65 } & \textbf{\scriptsize{}0.39 } & {\scriptsize{}0.48 } & {\scriptsize{}0.21 } & {\scriptsize{}0.95 } & \textbf{\scriptsize{}0.90 } & {\scriptsize{}0.35 } & {\scriptsize{}0.47 } & \textbf{\scriptsize{}0.33 }\tabularnewline
{\scriptsize{}VD2D3D\_v3} & {\scriptsize{}DenseCRF 2D } & {\scriptsize{}1.22 } & \textbf{\scriptsize{}0.18}{\scriptsize{} } & \textbf{\scriptsize{}0.21 } & \textbf{\scriptsize{}0.25 } & {\scriptsize{}0.68 } & {\scriptsize{}0.69 } & {\scriptsize{}0.48 } & \textbf{\scriptsize{}0.43}{\scriptsize{} } & \textbf{\scriptsize{}0.18 } & \textbf{\scriptsize{}0.94 } & {\scriptsize{}0.96 } & \textbf{\scriptsize{}0.29}{\scriptsize{} } & \textbf{\scriptsize{}0.46 } & {\scriptsize{}0.36 }\tabularnewline
\end{tabular*}
\end{table*}

\begin{table*}
\caption{Results of VD2D3D architecture using the DenseCRF 2D for segmentation,
with different metrics. The best measure (the lowest value) for each
individual stack and for statistical value is shown in bold. \label{tab:Results-of-VD2D3D}}

{\scriptsize{}}%
\begin{tabular*}{2\columnwidth}{@{\extracolsep{\fill}}>{\centering}p{0.2\columnwidth}|>{\centering}b{0.05\columnwidth}>{\centering}b{0.05\columnwidth}
>{\centering}b{0.05\columnwidth}>{\centering}b{0.05\columnwidth}
>{\centering}b{0.05\columnwidth}>{\centering}b{0.05\columnwidth}
>{\centering}b{0.05\columnwidth}>{\centering}b{0.05\columnwidth}
>{\centering}b{0.05\columnwidth}>{\centering}b{0.05\columnwidth}
>{\centering}b{0.05\columnwidth}>{\centering}b{0.05\columnwidth}|>{\centering}b{0.05\columnwidth}>{\centering}b{0.05\columnwidth}}
\textbf{\scriptsize{}Metric} & \textbf{\scriptsize{}1 } & \textbf{\scriptsize{}2 } & \textbf{\scriptsize{}3 } & \textbf{\scriptsize{}4 } & \textbf{\scriptsize{}5 } & \textbf{\scriptsize{}6 } & \textbf{\scriptsize{}7 } & \textbf{\scriptsize{}8 } & \textbf{\scriptsize{}9 } & \textbf{\scriptsize{}10 } & \textbf{\scriptsize{}11 } & \textbf{\scriptsize{}12 } & \textbf{\scriptsize{}Mean } & \textbf{\scriptsize{}SD }\tabularnewline
\hline 
{\scriptsize{}AUC } & {\scriptsize{}0.92 } & {\scriptsize{}0.93 } & {\scriptsize{}0.92 } & {\scriptsize{}0.89 } & {\scriptsize{}0.95 } & {\scriptsize{}0.96 } & {\scriptsize{}0.94 } & {\scriptsize{}0.95 } & {\scriptsize{}0.91 } & {\scriptsize{}0.94 } & {\scriptsize{}0.89 } & {\scriptsize{}0.94 } & {\scriptsize{}0.93 } & {\scriptsize{}0.02 }\tabularnewline
{\scriptsize{}ADJRIND } & {\scriptsize{}0.55 } & {\scriptsize{}0.76 } & {\scriptsize{}0.74 } & {\scriptsize{}0.64 } & {\scriptsize{}0.45 } & {\scriptsize{}0.50 } & {\scriptsize{}0.69 } & {\scriptsize{}0.68 } & {\scriptsize{}0.73 } & {\scriptsize{}0.58 } & {\scriptsize{}0.54 } & {\scriptsize{}0.70 } & {\scriptsize{}0.54 } & {\scriptsize{}0.19 }\tabularnewline
{\scriptsize{}MUTINF } & {\scriptsize{}0.28 } & {\scriptsize{}0.56 } & {\scriptsize{}0.62 } & {\scriptsize{}0.48 } & {\scriptsize{}0.11 } & {\scriptsize{}0.13 } & {\scriptsize{}0.27 } & {\scriptsize{}0.27 } & {\scriptsize{}0.55 } & {\scriptsize{}0.29 } & {\scriptsize{}0.38 } & {\scriptsize{}0.36 } & {\scriptsize{}0.31 } & {\scriptsize{}0.18 }\tabularnewline
{\scriptsize{}HDRFDST } & {\scriptsize{}47.05 } & {\scriptsize{}33.38 } & {\scriptsize{}82.76 } & {\scriptsize{}24.72 } & {\scriptsize{}35.37 } & {\scriptsize{}62.51 } & {\scriptsize{}26.87 } & {\scriptsize{}29.46 } & {\scriptsize{}23.45 } & {\scriptsize{}59.92 } & {\scriptsize{}73.12 } & {\scriptsize{}27.66 } & {\scriptsize{}37.59 } & {\scriptsize{}22.35 }\tabularnewline
\textbf{\scriptsize{}AVGDIST } & \textbf{\scriptsize{}1.44 } & \textbf{\scriptsize{}0.19 } & \textbf{\scriptsize{}0.23 } & \textbf{\scriptsize{}0.29 } & \textbf{\scriptsize{}0.67 } & \textbf{\scriptsize{}0.67 } & \textbf{\scriptsize{}0.48 } & \textbf{\scriptsize{}0.49 } & \textbf{\scriptsize{}0.18 } & \textbf{\scriptsize{}0.98 } & \textbf{\scriptsize{}0.98 } & \textbf{\scriptsize{}0.31 } & \textbf{\scriptsize{}0.49 } & \textbf{\scriptsize{}0.39 }\tabularnewline
{\scriptsize{}MAHLNBS } & {\scriptsize{}0.28 } & {\scriptsize{}0.06 } & {\scriptsize{}0.07 } & {\scriptsize{}0.15 } & {\scriptsize{}0.18 } & {\scriptsize{}0.13 } & {\scriptsize{}0.16 } & {\scriptsize{}0.03 } & {\scriptsize{}0.08 } & {\scriptsize{}0.03 } & {\scriptsize{}0.17 } & {\scriptsize{}0.08 } & {\scriptsize{}0.10 } & {\scriptsize{}0.07 }\tabularnewline
\end{tabular*}{\scriptsize{}}\\
{\scriptsize \par}

\scriptsize{AUC - Area Under the Curve, ADJRIND - Adjust Rand Index
considering a correction for chance, MUTINF - Mutual information,
HDRFDST - Hausdorff distance with the 0.95 quantile method, AVGDIST
- Average Hausdorff Distance, MAHLNBS - Mahalanobis Distance.}
\end{table*}

\begin{figure*}
\includegraphics[width=2\columnwidth]{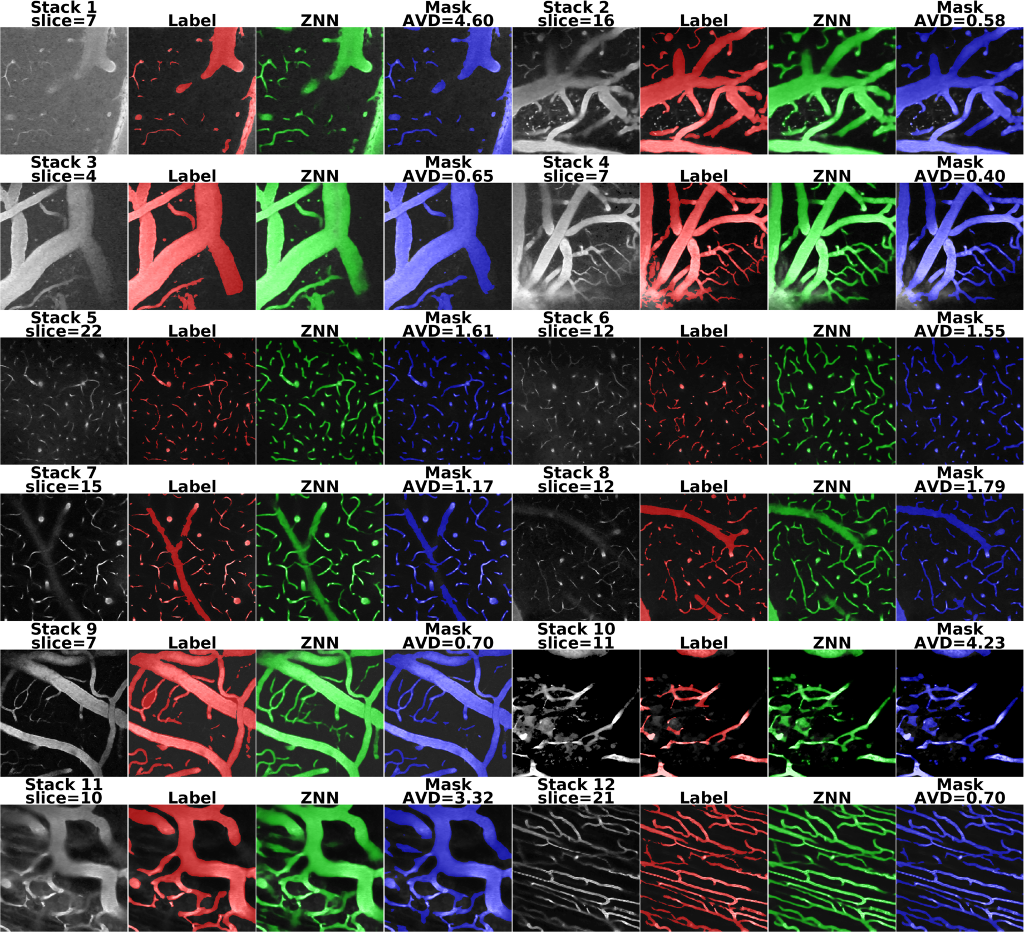}

\caption{\textbf{VD2D3D. Best} correspondence for each stack as evaluated by
Average Hausdorff distance. Architecture here VD2D3D, and segmentation
with dense CRF.\textbf{ \label{fig:Best-correspondence-for}}}
\end{figure*}

\begin{figure*}
\includegraphics[width=2\columnwidth]{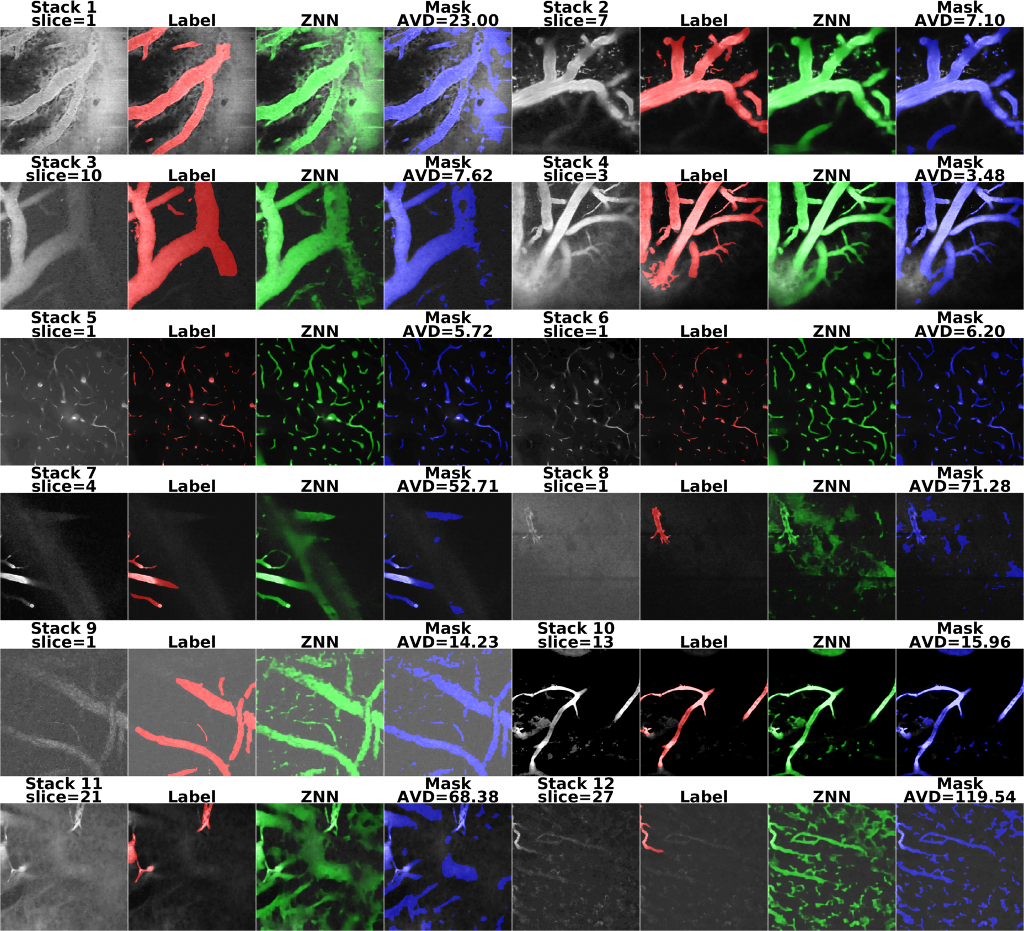}

\caption{\textbf{VD2D3D. Worst} correspondence for each stack as evaluated
by Average Hausdorff distance. The Stack 10 had erroneous correspondences
between the ground truth and the actual image explaining now the poor
performance. One could argue though that the results are not that
horrible, ZNN has found some faint vessels which are not labeled in
the ground truth at all. Architecture here VD2D3D, and segmentation
with dense CRF. \textbf{\label{fig:Worst-correspondence-for}}}
\end{figure*}

The segmentation results are visualized for the best slice for each
stack in \figref{Best-correspondence-for}, and for the worst slice
for each stack in \figref{Worst-correspondence-for}. For each stack
there are four columns: 1) the first column shows the denoised input
slice, 2) Label that corresponds to the manually annotated vessels,
3) the real-valued ZNN output from the proposed architecture, 4) the
Mask that is a binary mask obtained with dense two-dimensional CRF.
It should be noted that the ground truth labels are not optimally
defined, as can be seen for example in the worst case scenario of
stack \#3 (\figref{Worst-correspondence-for}) with high AVD value,
but visually the segmentation seems quite good. The high value of
AVD value simply comes from the difference between the suboptimal
manual label and the ``real'' vasculature labels that could have
been drawn better. 

Visualized segmentation results and the performance metrics for 
other VD2D3D variants are shown in the Wiki of our Github repository at \href{https://github.com/petteriTeikari/vesselNN/wiki}{https://github.com/petteriTeikari/vesselNN/wiki}.

\begin{figure}
\includegraphics[width=1\columnwidth]{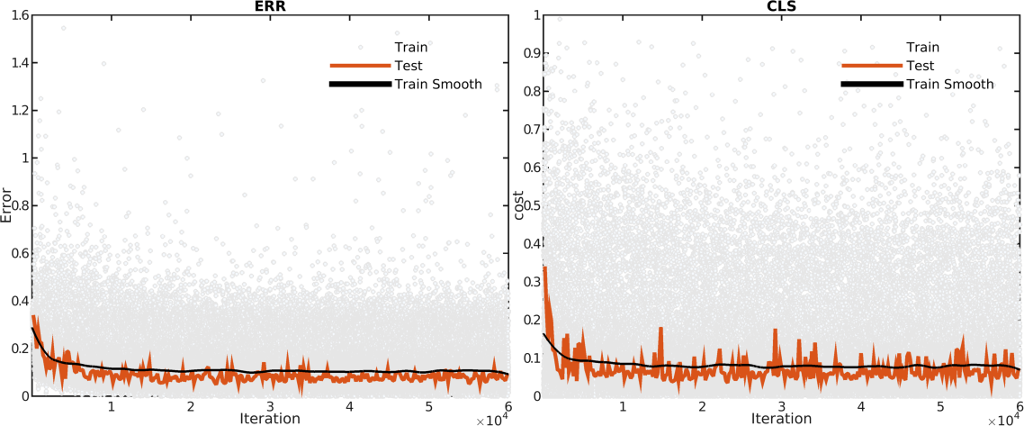}

\caption{Behavior of training and test error during training of \textbf{VD2D
architecture} (the first 60,000 iterations). ERR - Cost energy. CLS
- pixel classification error.\textbf{\label{fig:Behavior-of-training-VD2D}}}
\end{figure}

\begin{figure}
\includegraphics[width=1\columnwidth]{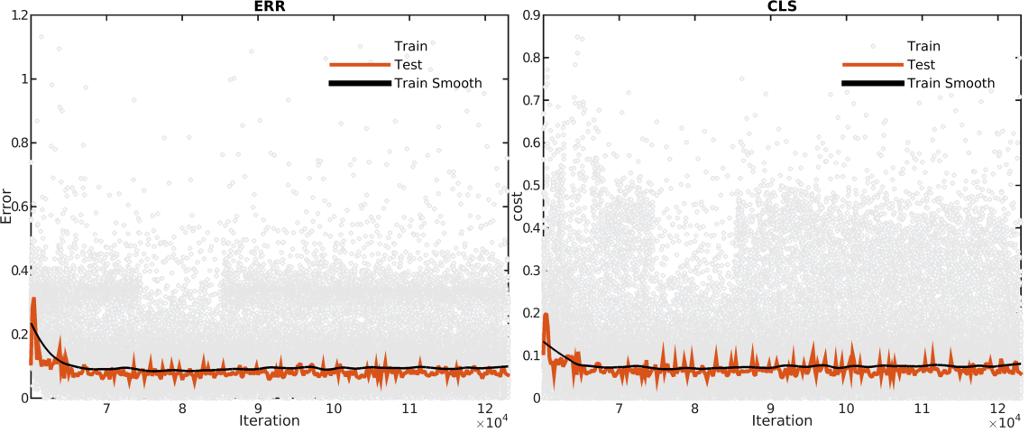}

\caption{Behavior of training and test error during training of \textbf{VD2D3D
architecture} (after the initial 60,000 iterations with the VD2D).
ERR - Cost energy. CLS - pixel classification error.\label{fig:Behavior-of-training-VD2D3D}}
\end{figure}

Visualization of the behavior of the network training for VD2D (\figref{Behavior-of-training-VD2D})
and for VD2D3D (\figref{Behavior-of-training-VD2D3D}) show that for
our datasets the training error (accuracy) and the test error (if
too high with low training error, the system is overfitting the training
data) converged well before the hard-coded limits taken from the study
of Lee \emph{et al. }\cite{lee2015recursive} for electron microscopy
stacks.

\section{Discussion}

Our proposed networks based on the ZNN framework \cite{lee2015recursive,zlateski2015imagesegmentation}
for vasculature segmentation from volumetric two-photon microscope
stacks provided promising results of segmentation quality. There is
still room for many improvements and optimizations to our proof-of-concept
approach which are discussed in more detail below.

\subsection{Deep learning}

\paragraph*{Refinements to network}

\label{sub:Refinements-to-networks}In this work, we chose to use
the ``vanilla'' network architecture from Lee \emph{et al.} \cite{lee2015recursive}
termed VD2D3D (``Very Deep 2D-3D'') with 2D layers in the initial
layers, and 3D layers at higher abstraction layers to make the network
faster to run and train. The VD2D3D employed commonly used components
of ConvNets with mixed nonlinear activation functions of hyperbolic
tangent ($tanh$) and rectified linear units (ReLU), and maximum filtering
variant of max pooling that kept the resolution the same throughout
the architecture without any need for upsampling as needed for some
architectures (e.g. Ronneberger \emph{et al.} \cite{ronneberger2015unetconvolutional}
for biomedical image segmentation).

The whole field of deep learning and ConvNets is rapidly advancing
(see for example a recent review by Gu \emph{et al.} \cite{gu2015recentadvances}).
We can thus expect that with future optimization and testing, the
``vanilla'' network can be improved for our application and for
volumetric biomedical segmentation in general. For example the convolutional
layers used now can be regarded as a generalized linear model (GLM)
for the the underlying local image patch, and the nonlinear learning
is introduced to the network via nonlinear activation function such
as Rectified Linear Units (ReLU). It has been proposed that the convolutional
filter itself could be made nonlinear with ``Network in Network''
(NIN) model of Lin \emph{et al.} \cite{lin2013network} or with the
Inception module by Szegedy \emph{et al.}\cite{szegedy2014goingdeeper,szegedy2015rethinking}.
These modifications enhance the abstraction ability of the local model
compared to the current GLM convolution model.

Very recently there has been interesting work of replacing convolutional
filter with bilateral filter \cite{kiefel2014permutohedral,gadde2015superpixel,jampani2015learning,barron2015thefast}
that is very commonly used edge-preserving smoothing filter \cite{tomasi1998bilateral}.
The convolutional filters were replaced both from earlier layers \cite{kiefel2014permutohedral},
as well as from later fully-connected layers \cite{gadde2015superpixel}
offering faster runtime especially for higher-dimensional signals.
Gadde \emph{et al.} \cite{gadde2015superpixel} replaced the Inception
modules with ``bilateral Inception'' superpixels yielding better
segmentation results than strictly pixel-wise implementations. Bilateral
Inception allowed long-range edge-preserving inference directly removing
the need for dense CRF as post-processing step according to the authors
\cite{gadde2015superpixel}. In contrast, Jampani \emph{et al.} \cite{jampani2015learning}
trained the bilateral filter to be used within the dense CRF inference,
demonstrating better segmentation performance compared to traditional
dense CRF. In general, introducing bilateral filter or some other
image-adaptive kernel at the convolutional layer level should allow
better edge-preserving properties of the network that is very useful
when we are interested in segmenting the vessel boundaries.

There have been many attempts to improve the max-pooling \cite{gu2015recentadvances}
of which the maximum filtering used here is a dense variant that retains
original volume resolution. Pooling in general is used to lower the
computation burden by reducing connections between successive layers.
From the recent efforts, especially spectral pooling seems like an
interesting upgrade \cite{rippel2015spectral} as it can be implemented
with little computational cost for Fast Fourier Transform (FFT) based
convolution networks such as the VD2D3D used here. In contrast to
max-pooling, the information is reduced in frequency domain in linear
low-pass filter fashion that will retain more information for the
same output dimensionality. The use of spectral pooling provided the
best classification performance on CIFAR (10 and 100) image classification
dataset \cite{krizhevsky2009learning} compared to other state-of-the-art
methods such as stochastic pooling\cite{zeiler2013stochastic}, Maxout
\cite{goodfellow2013maxoutnetworks}, ``Network in Network'' (NIN)
\cite{lin2013network}, and deeply-supervised nets \cite{lee2015recursive}.

Similarly, the traditional nonlinear activation functions such as
sigmoid, $tanh$\emph{, }and ReLUs could be improved. ReLUs are probably
the most commonly used activation function in ConvNets \cite{nair2010rectified},
with their main disadvantage being that it has zero gradient when
the unit is not active. This in practice may cause that the units
are not initially active never will become active during the gradient-based
optimization (stochastic gradient descent, SDG). To alleviate this
problem, Clevert \emph{et al.} \cite{clevert2015fastand} recently
proposed exponential linear units (ELUs) which also employ negative
values unlike ReLU, and according to the authors the use of ELUs lead
not only to faster learning, but also give better generalization performance
especially when the networks have at least 5 layers. On CIFAR-100
dataset, the ELUs yielded the best published result. The use of ELUs
would be in theory complimentary to spectral pooling and they could
also be used together with the nonlinear modifications of convolution
layer (e.g. NIN and Inception). It should be noted that at the moment
there is no nonlinear activation function for frequency domain \cite{rippel2015spectral},
thus there is a computational bottleneck with the inverse FFT and
FFT transforms needed before and after the activation function.

We employed Dropout \cite{srivastava2014dropout} for regularization
of our network by applying it before the output layer. Recently, Poole
\emph{et al.} \cite{poole2014analyzing} showed that injecting Gaussian
noise instead of applying Dropout led to improved performance, and
Rasmus \emph{et al.} \cite{rasmus2015semisupervised} found no practical
difference between Dropout and Gaussian noise injection. Interestingly
for Dropout, Gal and Ghahramani \cite{gal2015dropout}; and Kingma
\emph{et al.} \cite{kingma2015variational} demonstrated how deep
learning network with Dropout can be cast as a Bayesian model. This
in practice allows the estimation uncertainty based on Bayesian statistics
\cite{ghahramani2015probabilistic}. The estimate of uncertainty is
currently lacking in most of the deep learning frameworks. The advantage
of the Dropout-based Bayesian estimation is that one can turn existing
dropout networks to include model uncertainty, rather than having
to re-define the whole architecture. This Dropout-based estimation
was used by Kendall \emph{et al.} \cite{kendall2015bayesian} for
semantic segmentation showing comparable performance to state-of the-art
architectures by applying Dropout in the central layers of their encoder-decoder
architecture. In analysis pipelines where a quantitative analysis
of morphological vessel behavior (e.g. \cite{lindvere2013cerebral})
follows the image processing, it is useful to propagate the uncertainties
involved in the image processing pipeline to the final statistical
analysis.

The most obvious improvement for the used VD2D3D architecture here
would be the conversion of all the convolutional layers to be three-dimensional.
However, this is not computationally that feasible using current ZNN
implementation with most commonly available hardware around. In the
future with increased computational power, and speed optimization
this should become feasible either by using Intel Xeon coprocessor
\cite{rippel2015spectral,zlateski2015imagesegmentation}, supercomputing
clusters \cite{zhang2015areliable}, or GPU-accelerated frameworks
such as Theano \cite{thetheanodevelopmentteam2016theanoa}. In our
current implementation we chose to do the dense CRF in slice-by-slice
manner due to the available implementation of it. In the future, we
could upgrade the used dense CRF to three dimension as done for example
by Kundu \emph{et al. }\cite{kundu2016feature}.

In the architecture employed here, multi-scale representation is not
explicitly included. We have tried to provide stacks with different
magnifications in our dataset to help the network learn different
scales like done by Lee \emph{et al.} \cite{lee2015recursive}. Typically
in semantic segmentation networks, multi-scale representation is implemented
in two main ways \cite{chen2015attention}, either by using so called
\emph{skip-net }that combine features from the intermediate layers
of network \cite{sermanet2013overfeat,chen2015semantic,long2014fullyconvolutional},
or via \emph{share-net }that are fed input resized to different scales
\cite{lin2015efficient,farabet2013learning}. The discussed bilateral
filter modification would be able to encode scale invariance defined
on continuous range of image scales without the typically used finite
number of subsampled inputs simplifying the network architecture \cite{kiefel2014permutohedral}.

In addition to concentrating on the individual components of the ConvNets,
there have been alternative approaches to improve computational efficiency
\cite{cheng2015anexploration,zhang2015supervised,ioffe2015batchnormalization,gupta2015modelaccuracy}.
Our vessel segmentation network took over 20 days (see \ref{sub:Training-procedure})
to train on a typical multicore desktop computer, which emphasizes
the utility of faster computation. Batch Normalization technique
by Ioffe \emph{et al.} \cite{ioffe2015batchnormalization} has received
a lot of attention as the authors showed that the same classification
accuracy can be obtained with 14 times fewer training steps while
exceeding accuracy of human raters with an ensemble of batch-normalized
networks. By normalizing for each training mini-batch, higher learning
rates could be used with the training being less sensitive to initialization
as well.

Another typically used speedup scheme is to use superpixels \cite{nunez-iglesias2013machine,farag2015abottomup,gadde2015superpixel}
with two-dimensional images, or supervoxels \cite{lucchi2012supervoxelbased,konyushkova2015introducing}
with volumetric three-dimensional images to reduce the dimensionality
of the input. Within the superpixel/supervoxel pixels/voxels carry
similarities in color, texture, intensity, etc., generally aligning
with region edges, and their shapes being generally circular/spherical
rather than rectangular patches. The main downside of superpixels/supervoxels
are that they introduce a quantization error \cite{gadde2015superpixel}
whenever pixels/voxels within one segment have different ground truth
label assignments (i.e. in our case supervoxel would have both non-vessel
and vessel labels).

One of the main bottlenecks currently in deep learning networks, is
the lack of efficient algorithms and libraries for sparse data, as
majority of the libraries are optimized for dense data \cite{szegedy2014goingdeeper}.
The already discussed introduction of bilateral filters, and their
computation using permutohedral lattices \cite{adams2010fasthighdimensional,kiefel2014permutohedral}
is a one way to speedup the computation of sparse data. In addition
to permutohedral lattice, Ghesu \emph{et al.} \cite{ghesu2015marginal}
introduced a Marginal Space Deep Learning (MSDL) framework for segmenting
volumetric medical images by replacing the standard, pre-determined
feature sampling pattern with a sparse, adaptive, self-learned pattern
showing increased runtime efficiency.

\paragraph*{Improved annotation\label{sub:Improved-annotation}}

We manually annotated our ground truths using Matlab-created seeds
and GIMP (GNU Image Manipulation Program). This was extremely time-consuming
and required a person familiar with the two-photon microscopy vasculature
images. Recently Mosinska \emph{et al.} \cite{mosinska2015activelearning}
extended the active learning (AL) approach (\cite{settles2010activelearning})
for delineation of curvilinear structures including blood vessels.
Active learning is designed to reduce the effort of the manual annotator
by selecting from non-annotated dataset, the image stacks for manual
annotation that would the most beneficial for improving the performance
of the network. Surprisingly and counter-intuitively, recent work
on electron microscope image segmentation \cite{konyushkova2015introducing}
found that the classifier performance of their implementation was
better using only a subset of the training data instead of using the
whole available training data. This phenomenon had been reported before
by \cite{schohn2000lessis}, suggesting that a well chosen subset
of training data can produce better generalization than the complete
set.

\paragraph*{Crowdsourcing}

Kim \emph{et al.} \foreignlanguage{american}{\cite{kim2014spacetime}}
demonstrate an interesting approach for acquiring annotations for
electron microscopy datasets by developing a game called EyeWire (\href{http://eyewire.org/}{http://eyewire.org/})
for non-experts where they can solve spatial puzzles made out from
neuronal boundaries. This crowdsourcing have been traditionally used
in tasks that does not require expert-level knowledge such as teaching
autonomous cars to drive \foreignlanguage{american}{\cite{rajpurkar2015driverseat}},
but have been thought to be impractical for tasks that require expertise
such as medical segmentation \foreignlanguage{american}{\cite{mosinska2015activelearning}}.
The innovative approach used in their game is able to transform the
biomedical ``expert'' annotation problem to the masses. 

Additionally to the ``gamification'' of segmentation efforts, one
could create a segmentation challenge of our dataset to popular machine
learning sites such as Kaggle (\href{https://www.kaggle.com/}{https://www.kaggle.com/})
and Grand Challenges in Biomedical Analysis (\href{http://grand-challenge.org/}{http://grand-challenge.org/})
to bring up the volumetric vascular segmentation in par with the rest
of biomedical image analysis domains with existing datasets.

\subsubsection*{Unsupervised pre-training}

Another way to reduce the labor-intensive ground truth annotation
required for our supervised approach, would be to initialize our supervised
network using unsupervised pre-training from non-annotated dataset
(\cite{bengio2007greedylayerwise}). In practice, we would feed the
unsupervised learning network all our existing vascular image stacks
without any annotation labels, and the network would learn the most
representative features of that dataset that could be then fed into
the first layer of our supervised network (\texttt{Conv1a} of \figref{An-overview-of-ZNN-architecture}).
Erhan \emph{et al. }\cite{erhan2010whydoes} have suggested that this
pre-training initialization serves as a kind of regularization mechanism
that is retained even during the supervised part with the classification
performance not deteriorating with the additional supervised training.
We could for example use the dictionary learning approach with sparsity
priors for 2D vessel images and 3D neuron dendrites proposed by \cite{sironi2015learning}
as the pre-processing step, or alternatively use some stacked autoencoder
variant used for medical image segmentation \cite{shin2013stacked,suk2015deeplearning}.

More elegant alternative for unsupervised pre-training is to simultaneously
apply both unsupervised and supervised learning, instead of having
unsupervised pre-training and supervised training as separate steps
\cite{rasmus2015semisupervised,maaloe2015improving}. Rasmus \emph{et
al.} \cite{rasmus2015semisupervised} proposed a modified Ladder Network
\cite{valpola2014fromneural} which demonstrate how by adding their
unsupervised Ladder Network to existing supervised learning methods
including convolutional networks improved significantly classification
performance in handwriting classification (MNIST database \cite{lecun1998gradientbased}),
and in image classification (CIFAR-10 database \cite{krizhevsky2009learning})
compared to previous state-of-the-art approaches. Their approach excelled
when the amount of labels were small, and especially when number of
free parameters was large compared to the number of available samples,
showing that the model was able to use the unsupervised learning part
efficiently. Particularly attractive detail of their publicly available
approach , is that it can be added relatively easy on a network originally
developed for supervised learning such as ours, allowing hopefully
a better use of our limited annotated dataset.

\subsubsection*{Joint training of the image processing pipeline}

\label{sub:Training-everywhere}In our work, we have only focused
on replacing the vessel enhancement step (see \figref{Typical-vessel-segmentation-pipeline})
with automated data-driven ConvNet assisted by various parametrized
filters requiring some degree of user interaction. Ideally we would
like to all relevant steps starting from image restoration to post-processing
of the volumetric ConvNet output, all the way to the mesh generation
to be automated using training data to increase the robustness and
minimize user interaction.

Work has already been done for each individual components that could
be simply stacked together as separate units, or one could jointly
train all components in end-to-end fashion. For example recent work
by Vemulapalli \emph{et al.} \cite{vemulapalli2015deepgaussian} showed
that their deep learning network based on a Gaussian Conditional Random
Field (GCRF) model outperformed existing methods in two-dimensional
image denoising including the two-dimensional variant BM3D \cite{dabov2007imagedenoising}
of the BM4D algorithm \cite{maggioni2013nonlocal} that we used to
denoise our vessel stacks. For other image restoration task such as
blind deconvolution \cite{xu2014deepconvolutional} for sharpening
the stacks, blind inpainting \cite{cai2015blindinpainting} for filling
possibly broken vessels, vibration-artifacts, or other image quality
artifacts, and motion-blur correction \cite{sun2015learning} deep
learning based solutions have been proposed with promising results.

Recent work by Xu \emph{et al.} \cite{xu2015deepedgeaware} demonstrate
a deep convolutional networks designed to learn blindly the output
of any deterministic filter or a combination of different filters.
Authors demonstrated this by learning two different edge-preserving
smoothing filters bilateral filter (\cite{tomasi1998bilateral,barron2015thefast}),
and $L0$ gradient minimization smoothing (\cite{xu2011imagesmoothing})
jointly without needing to know anything about the implementations
of such filters given that input and output images can be accessed.
This edge-aware smoothing could be used as a refining step for our
image denoising/deconvolution output to further suppress irrelevant
structure for the vessel segmentation. Alternatively, the same framework
could be potentially to learn the behavior of commercial software
as demonstrated by the authors with \emph{``copycat filter scheme''}
using Photoshop\textsuperscript{\textregistered} filters \cite{xu2015deepedgeaware}.
One could generate training data for deconvolution for example using
some commonly used software package such as Imaris (Bitplane AG, Zurich,
Switzerland) or AutoQuant (AutoQuant Imaging/Media Cybernetics), and
integrating that ``knowledge'' to the same deep learning framework
without having to jump between different software packages during
the analysis of microscopy stacks.

Lee \emph{et al.} \cite{lee2015recursive} argue that the recursive
input from VD2D can be viewed as modulatory 'gate' that the feature
activations for structures of interest are enhanced while suppressing
activations unrelated to structures of interest. Based on that assumption,
it would be interesting to try to replace the VD2D altogether for
example with data-driven edge detection network such as the N\textsuperscript{4}-fields
\cite{ganin2014ntextasciicircum4fields} or holistically-nested edge
detection \cite{xie2015holisticallynested}. N\textsuperscript{4}-fields
\cite{ganin2014ntextasciicircum4fields} was shown to segment two-dimension
retinal vasculature from the DRIVE dataset \cite{staal2004ridgebased}
better than the Structured Edge detector \cite{dollar2015fastedge}
while the performance was not compared to traditional vessel enhancement
filters. Alternatively one could try to integrate recent vessel enhancement
filters as structured layers \cite{ionescu2015matrixbackpropagation}
within the ConvNet architecture to try to incorporate some domain
knowledge without having to resort to totally hand-crafted features.
Recent vesselness filters of interest include the scale-invariant
enhancement filter by Moreno \emph{et al.} \cite{moreno2015gradientbased},
and the nearest neighbor-inspired detection of elongated structures
by Sironi \emph{et al.} \cite{sironi2015projection}.

The deep learning can be seen as a ``brute force'' method for vessel
segmentation as it does not explicitly model the geometrical relationships
that exist between neighboring ``vessel pixels'' as pointed out
by Sironi \emph{et al.} \cite{sironi2015projection}. The probability
maps can have isolated erroneous responses, discontinuities and topological
errors that are typically mitigated using post-processing techniques
such as Conditional Random Fields (CRF, \cite{krahenbuhl2012efficient,chen2015semantic,lin2015efficient}),
narrow-band level sets \cite{kohlberger2011automatic}, learned graph-cut
segmentation \cite{wolz2013automated} or Auto-Context \cite{tu2010autocontext}
among others. Authors of the ZNN framework \cite{lee2015recursive}
chose to refine their segmentation of electron microscope stacks using
a watershed-based algorithm developed by themselves \cite{zlateski2015imagesegmentation},
whereas recent work by Almasi \emph{et al.} \cite{almasi2015anovel}
reconstructed microvascular networks from the output of active contours
\cite{chan2001activecontours}, and Sironi \emph{et al.} \cite{sironi2015projection}
train an algorithm inspired by Nearest Neighbor Fields \cite{ganin2014ntextasciicircum4fields}
to induce global consistency for the probability maps. Both those
recent works \cite{almasi2015anovel,sironi2015projection} can be
seen complimentary and refining post-processing steps to our approach.

At the moment we are only training individual stacks at the time,
but it is common in biomedical microscopy to image the same stack
over multiple time points. We could extend our model to exploit the
temporal dependency among multiple time points, as it is done in 2D
video processing where the time can be regarded as the third dimension.
Huang \emph{et al.} \cite{huang2015bidirectional} for example employ
a recurrent neural network (RNN) for modeling temporal context in
a video sequence for multi-frame super-resolution reconstruction.
This potentially can improve the vessel segmentation as the vessels
are not typically deformed heavily between successive stacks when
using typical acquisition intervals. The time-extended super-resolution
approach should in theory improve the quality of the interpolation
in $z$- dimension when isotropic voxels are wanted, compared to deep
learning based single-frame super-resolution \cite{kim2015deeplyrecursive},
and traditional B-spline interpolation \cite{indhumathi2012adaptiveweighted}.

To the knowledge of the authors, there has been no attempt to improve
the mesh reconstruction step using deep learning framework. Closest
example to deep learning in surface reconstruction were demonstrated
by Xiong \emph{et al.} \cite{xiong2014robustsurface}, who used a
dictionary learning for surface reconstruction from a point cloud,
which outperformed state-of-the art methods in terms of accuracy,
robustness to noise and outliers, and geometric feature preservation
among other criteria. Jampani \emph{et al.} \cite{jampani2015learning}
demonstrated how they could learn optimal bilateral filter parameters
for three-dimensional mesh denoising that could be thus used as a
post-processing step for surface reconstruction. This is an improvement
of the bilateral filter mesh denoising algorithm implemented in Computational
Geometry Algorithms Library (CGAL, \href{http://www.cgal.org/}{http://www.cgal.org/})
that requires user-set parameters.

\begin{figure*}
\centerline{\includegraphics[width=1.5\columnwidth]{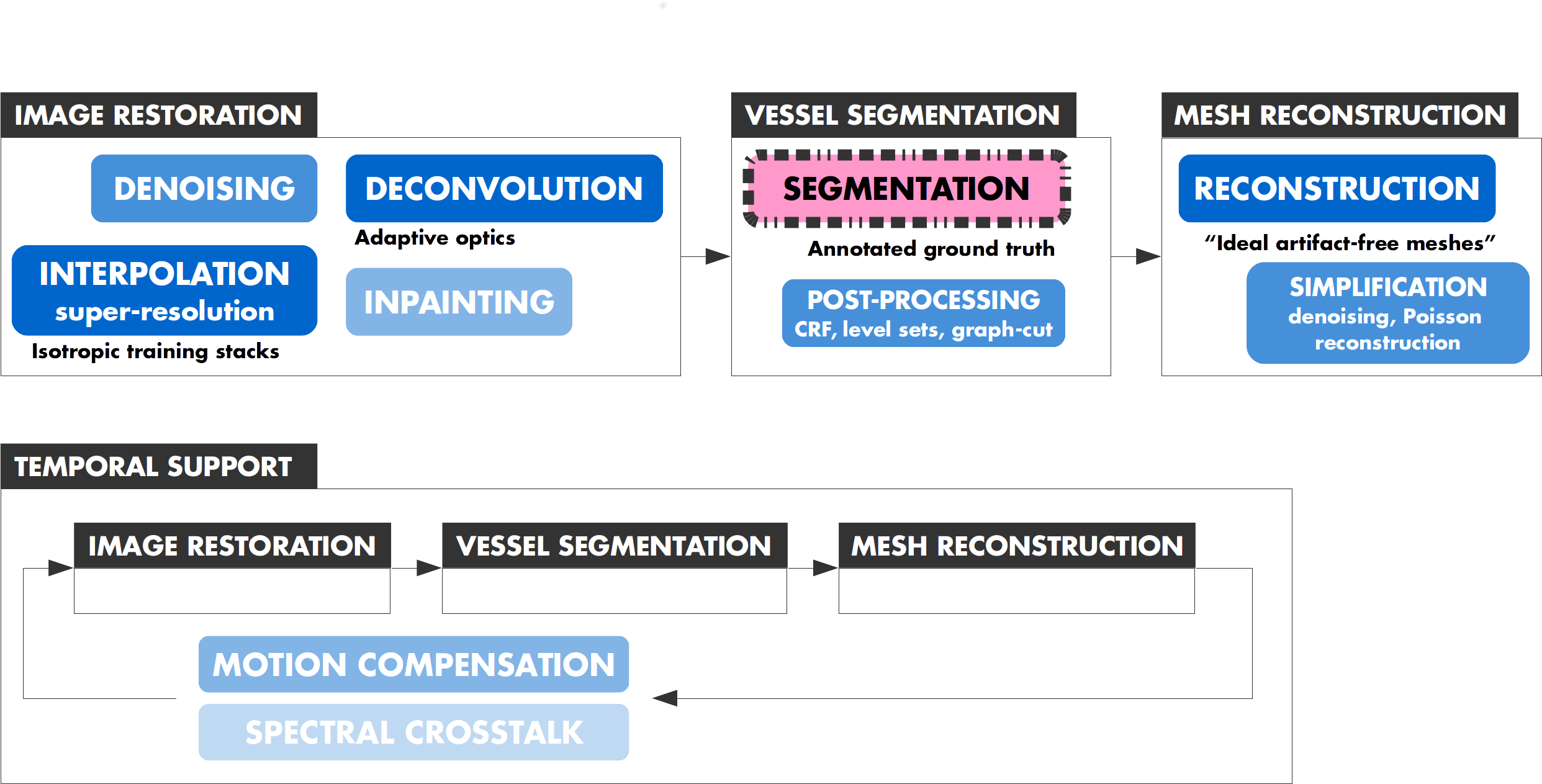}}

\caption{Example schematic of fully trainable pipeline for vascular segmentationk.
\textbf{(top) }Segmentation pipeline of a single stack.\textbf{ }The
pipeline is divided into three sub-components: image restoration,
vessel segmentation and mesh reconstruction. The image restoration
part could for example consist of joint model for denoising \cite{vemulapalli2015deepgaussian},
deconvolution \cite{xu2014deepconvolutional}, interpolation (super-resolution)
\cite{kim2015deeplyrecursive}, inpainting \cite{cai2015blindinpainting},
motion artifact correction, and image-based spectral unmixing if multiple
dyes were used. \textbf{}\protect \\
\textbf{(bottom) }Segmentation pipeline of a stack with multiple time
points. The added temporal support is needed to estimate motion artifacts
\cite{soulet2013automated}, and it is able exploit the temporal dependency
of vasculature (i.e. vascular diameter and position changes are not
dramatic, \emph{better phrase here maybe}) and in theory should improve
 the estimates of all the sub-components compared to the single stack
scheme, as for example is the case for super-resolution \cite{huang2015bidirectional}.
If multiple dyes are used simultaenously there is a potential problem
of the dye signals ``leaking'' to other spectral channels that need
to be mitigated computationally using for example some blind image
separation technique \cite{abolghasemi2012blindseparation}. The spectral
crosstalk correction could be done for a single stack, but here we
assumed that more input data would allow more robust estimation of
the mixed image sources (e.g. with fast independent component analysis
\cite{himberg2003icassosoftware} ). \textbf{\emph{}}\protect \\
\textbf{\emph{}}\label{fig:Simplified-end-to-end-pipeline}}
\end{figure*}

The simplified schematic of the components for joint optimization
is shown in \figref{Simplified-end-to-end-pipeline}. In our proposed
approach we have only focused on the segmentation part whereas in
optimal case we would like to have training data of all the different
phases of the image processing pipeline. The schematic does not show
any more sophisticated layers that could be embedded inside of more
generalistic convolutional networks. For example Ionescu \emph{et
al.} \cite{ionescu2015matrixbackpropagation} demonstrated how to
backpropagate global structured matrix computation such as normalized
cuts or higher-order pooling. The training of normalized cuts within
deep learning framework is similar to the approach taken bu Turaga
\emph{et al. }\cite{turaga2009maximin}\emph{ }for optimizing a Rand
index with simple connected component labeling (MALIS, which is to
be implemented in the ZNN framework used by us). Inclusion of such
global layers was shown to increase the segmentation performance compared
to more generalized deep networks.

\subsubsection*{Other libraries}

Currently there are not many publicly available software for dense
image segmentation for volumetric 3D data, so we were constrained
in our choice between GPU-accelerated Theano \cite{thetheanodevelopmentteam2016theanoa}
and the CPU-accelerated ZNN \cite{zlateski2015znn}. We chose to
use the ZNN framework for our vessel segmentation pipeline.\emph{
}The Caffe-derived DeepLab \cite{chen2015semantic,papandreou2015weaklyand}
with both CPU and GPU acceleration options was not supporting efficient
3D ConvNets as it were the case with Caffe itself \cite{jia2014caffeconvolutional}
as benchmarked by Jampani \emph{et al.} \cite{jampani2015learning}
for example.
 The CPU-accelerated ZNN was shown to have efficient computational
performance compared to GPU-accelerated Theano \cite{zlateski2015znn}, and 
considering the recent price drop of Intel Xeon Phi\textsuperscript{TM} Knights
Corner generation of CPU accelerator cards, and introduction of supposedly
more user-friendly Knights Landing generation, our choice of implementation
seems relatively easy to accelerate in the future. Recently published Python library
Keras (\href{http://keras.io/}{http://keras.io/}) functions as a
high level abstraction library for either Theano and for TensorFlow
\cite{rampasek2016tensorflow} so that the researcher can focus on
the ideas and change flexibly the underlying backend between Theano
and TensorFlow as one wishes.

\subsection{Connection to other software frameworks}

Our vessel segmentation pipeline essentially replaces the previously
used handcrafted vesselness filters (e.g. \cite{frangi1998multiscale,turetken2012automated,moreno2015gradientbased})
still requiring a refining segmentation algorithm for the ZNN output
as the output is not a binary-valued mask, but rather a real-valued
probability map. Sumbul \emph{et al. }\cite{sumbul2014automated}
used connected component clustering (\texttt{bwlabeln} of Matlab,
union-find algorithm, \cite{fiorio1996twolinear}) with morphological
filters to refine the ZNN output for retinal ganglion cell (RGC) arbor,
while the most recent paper with ZNN \cite{lee2015recursive} compared
clustering to more sophisticated watershed-based segmentation \cite{zlateski2015imagesegmentation}
for segmenting neuronal boundaries from EM stacks.

Our work can be seen also as a pre-processing step for morphological
reconstruction of vessel networks in mesh domain. The output from
our pipeline could be for example used as an input for the mesh reconstruction
pipeline of Python-based open source Vessel Modeling Toolkit (VMTK,
\href{http://www.vmtk.org/}{http://www.vmtk.org/}), and its inexpensive
graphical front-end VMTKLab (\href{http://vmtklab.orobix.com/}{http://vmtklab.orobix.com/}).
This would be more robust segmentation pre-processing step compared
to the ones provided by VMTK. VMTK provided the following four vesselness
enhancing filters: 1) Frangi's method \cite{frangi1998multiscale},
2) Sato's method \cite{sato19973dmultiscale}, 3) Vessel Enhancing
Diffusion Filter \cite{enquobahrie2007vesselenhancing}, and 4) Vessel
enhancing diffusion \cite{manniesing2006vesselenhancing}, with the
Frangi's method being the default option. Vessel enhancing filter
works as a pre-processing step in VMTK pipeline for the level set
based vessel segmentation of VMTK before running the Marching Cubes
algorithm derivative \cite{lorensen1987marching} for mesh reconstruction.

For researchers who are the most comfortable using graphical tools
such as Imaris (Bitplane AG, Zurich, Switzerland), or open-source
ImageJ/FIJI platform \cite{schindelin2015theimagej}, the proposed
approach can be seen as automatic pre-processing step improving the
performance of the following manual processing steps. For example
the two-class (vessels, and non-vessels) vessel segmentation in Imaris
by \cite{lindvere2013cerebral}, required many user-supplied intensity
thresholds which could have been automatized with our ConvNet-based
approach, and the remaining steps for graph reconstruction could have
done with existing pipeline.\textbf{\emph{}}

\subsection{2-PM/Microscopy specific suggestions}

In addition to optimizing our algorithm, image parameters should also
be carefully chosen to facilitate vessel segmentation. We are interested
in quantifying the degree of blood-brain barrier opening (BBBO) following
focused ultrasound stimulation \cite{cho2011twophoton,nhan2013drugdelivery,burgess2014analysis}.
Experimentally, this is achieved by injecting a fluorescent dextran
into the systemic vasculature, and then measuring the difference in
fluorescence intensity between the vessels (foreground) and the surrounding
tissue during BBBO \cite{nhan2013drugdelivery,burgess2014analysis,yoon2015invivo}. Thus, by the nature of the experiment, we are making the task
harder for the segmentation network as the edges between the vessels
and the background will become progressively blurred. 

One way to improve such a loss of contrast is to quantify the BBBO
by using two vascular dyes simultaneously, one which readily leaks
out from vessels upon BBBD, and another one with a high molecular
weight that leaks out less. An alternative to using high-molecular
weight dextrans is to use quantum dots that have narrower emission
spectra for reduced dye crosstalk \cite{wegner2015quantum}, and
less leakage from vessels. Quantum dots have already been used to
study the tumor vasculature \cite{stroh2005quantum}. Another option
is to use Alexa Fluor 633 dye, which selectively labels the walls
of arteries that are greater than 15-$\mu$m in diameter\cite{shen2012anarteryspecific}.
This would make vessel segmentation easier as the 'leakage' channel
(with the dextran) and 'vessel' channel (with the labeled vessel walls)
can be analyzed separately. Recently, multiphoton fluorescent dyes
with longer emission and excitation wavelengths \cite{oheim2014newredfluorescent,kim2015twophoton}
have been gaining popularity due to their better transmission through
biological tissue yielding  improved penetration depths and signal-to-noise
ratios (SNRs) \cite{smith2009bioimaging,horton2013invivo}.

Another promising, yet not commonly employed, technique is to increase
excitation laser wavelengths up to 1,700 nm \cite{horton2013invivo},
and switch to three-photon excitation. This also improves depth penetration,
but also allows better optical sectioning due to higher non-linearity
due to the $z^{4}$ attenuation from the focal plane instead of $z^{2}$
attenuation in two-photon regime, where $z$ is the distance \cite{horton2013invivo}.
This reduces noise from out-of-planes and tissue autofluorescence
\cite{blab2001twophoton}. In terms of our future versions of deep
learning framework, we would like to simultaneously dyes for both
two-photon and three-photon process so that the crosstalk in $z$-dimension
would be minimized for three-photon process dye allowing that to be
used as the ground truth for the super-resolution training (see \figref{Simplified-end-to-end-pipeline})
for two-photon process dyes. Likewise the improved SNR either with
longer-wavelength dye and/or three-photon microscopy could be used
as the ground truth for the denoising block for denoising shorter-wavelength
fluorescent dyes.

Another way to improve SNR is to correct the optical aberrations caused
by brain tissue in real-time by using adaptive optics \cite{booth2014adaptive}.
The use of adaptive optics originated from astronomy \cite{babcock1953thepossibility},
where the correction of aberrations caused by atmospheric was able
to give better image quality to astronomers. Ji \emph{et al. } \cite{ji2012characterization}
demonstrated the increase in SNR for \emph{in vivo }calcium imaging
was especially significant atgreater depths. The better image quality
with adaptive optics could be used as the ground truth for the deconvolution
block (see \figref{Simplified-end-to-end-pipeline}) and the stack
without adaptive optics as the training data. Ideally, one could combine
all the above methods for optimized imaging quality.

\subsubsection*{Physiological refinement}

In the proposed architecture here, we did not explicitly try to further
refine the segmented vasculature to subclasses, but rather simply
differentiated vessel and non-vessel voxels. There have been some
work devoted to separating arteries from veins either using computational
techniques \cite{mehrabian2012aconstrained,estrada2015retinal}, or
using specific fluorescent labels that specifically label arteries
such as Alexa Fluor 633 used by Shen \emph{et al. }\cite{shen2012anarteryspecific}.
In the future, we would like extend our network to differentiate arteries
from veins by acquiring training data using such an artery-specific
dye concurrently with a fluorescent dextran that would label the entire
vascular network.

\subsubsection*{Extension to other medical applications}

In our ``vanilla network'' (see \ref{sub:Refinements-to-networks})
we did not have any vasculature specific optimization, and we decided
to leverage on the ability of deep learning network to learn the relevant
features itself of relying on handcrafted features. Thus, the same
network proposed initially for electron microscope image segmentation
\cite{sumbul2014automated,lee2015recursive} can be extended to other
applications as demonstrated here for volumetric two-photon vasculature
image segmentation. To extend the framework for other application,
annotated training data is needed for training the network for the
given task. To be used with vasculature datasets such as the VESSEL12
\cite{rudyanto2014comparing}, it would be sufficient to use our pre-trained
network and fine-tune the model, training with small learning rate,
rather having to learn from scratch as typically done in specific
image classification tasks exploiting some pre-trained network with
broader dataset \cite{carneiro2015unregistered,zhang2015deepmodel}.
This is known as transfer learning or domain adaptation depending
on the marginal data distribution \cite{patricia2014learning}. 

In practice with vascular segmentation, transfer learning approach
correspond to a situation when a network trained for tubular dataset
such as DIADEM \cite{brown2011thediadem,peng2015fromdiadem} is used
as a basis, and fine-tuning that network using limited samples of
multiphoton microscopy data. Domain adaptation would correspond to
a situation where we would have trained our network to segment vasculature
using some other imaging modality than multiphoton microscopy in which
the vasculature (foreground) itself might have similar appearance
to multiphoton microscopy, but the background from which we try to
segment the vasculature would be different. Xie \emph{et al.} \cite{xie2015hybridcnn}
combined ConvNet with a traditional dictionary-learning approach for
domain adaptation that was able to exploit the local discriminative
and structural information more efficiently than just using a ConvNet.
This is of a relevance for us, as we could use the unsupervised dictionary-based
learning approach for vessel stacks proposed by Sironi \emph{et al.
}\cite{sironi2015learning}, and combine that to our ConvNet-based
approach to exploit the large number of unlabeled vessel stacks.\textbf{}

In medical applications, there have been some effort of going around
the high annotation cost by exploiting auxiliary data such as textual
reports \foreignlanguage{american}{\cite{schlegl2015predicting,shin2015interleaved},
or image-level labels} \cite{kraus2015classifying} (i.e. whether
the whole stack/slice image contains a vessel or not). This type of
learning is known as weakly-supervised segmentation, and cannot understandingly
reach the segmentation performance as full pixel-level ``strong''
annotated supervised learning. Hong \emph{et al.} \cite{hong2015learning}
recently demonstrated that the gap between fully supervised and weakly-supervised
can be reduced compared to previous approaches by exploiting pre-trained
ImageNet model for transfer learning with weak labels. In multiphoton
microscopy, it is not typically possible to use whole-image labels
as the vasculature is typically so dense that there are not a lot
of empty slices with no vessel labels.. Sometimes in practice, the
dye loading is unsuccessful or there are technical glitches, and these
empty acquired empty stacks could be used to characterize the noise
characteristics of non-vessel areas.

\subsection{Open-source code, reproducibility}

\label{sub:Open-source-code,-reproducabilit}We share our annotated
two-photon vasculature dataset to the scientific community to address
the lack of standardized datasets for multiphoton microscopy. We believe
that part of the reason for lack of published work on volumetric vessel
segmentation is due to lack of suitable training data, most of the
biomedical image segmentation efforts being directed to fields such
as electron microscopy \foreignlanguage{american}{\cite{wu2015aniterative,lee2015recursive,maitin-shepard2015combinatorial,ronneberger2015unetconvolutional}},
and various clinical applications \foreignlanguage{american}{\cite{havaei2015braintumor,stollenga2015parallel,schlegl2015predicting,dubrovina2015computational}}
as the training data is readily available. We want to be part of creating
a cultural shift from independent efforts of research groups toward
an open source and collaborative neuroscience as datasets get larger
and more complex \cite{freeman2015opensource,gao2015onsimplicity},
as well as ensuring that our framework can be easily reproduced and
developed further \cite{piccolo2015toolsand}. In the future, we would
like to move away from proprietary Matlab environment to totally open-source
code in Python as well.

\section{Conclusion}

We have a proposed a deep learning based framework for two-class segmentation
(vessel, and non-vessel) of vascular networks obtained via two-photon
microscopy from mouse cortex and human squamous cell carcinoma tumors.\emph{
}We have made the Matlab code available based on the open-source ZNN
framework \cite{lee2015recursive,zlateski2015znn}. In contrast to
GPU-accelerated frameworks such as Theano \cite{thetheanodevelopmentteam2016theanoa},
the ZNN is optimized to run on CPU while reaching relatively similar
performance compared GPU-accelerated approaches \cite{zlateski2015znn}.
We have already made our training set freely available to address
the lack of annotated reference dataset for multiphoton microscopy
vasculature segmentation. We are hoping that this will both inspire
other research groups sharing their vasculature datasets, as well
as improving our proposed framework. Our future work will focus on
enhancing the computational performance and accuracy of the network
for multiphoton microscopy vessel segmentation.

\subsection*{Acknowledgements}

We would like to thank Sharan Sankar for his work as a summer student
writing wrapper for various wrappers for ITK C++ functions.

\subsection*{}

\scriptsize{
\bibliographystyle{plainurl_PT}
\addcontentsline{toc}{section}{\refname}\bibliography{../2pmNeuroMod}
}

\end{document}